\documentclass[11pt]{article}

\usepackage[preprint]{acl}

\usepackage{times}

\usepackage{latexsym}
\usepackage[T1]{fontenc}
\usepackage[utf8]{inputenc}
\usepackage{microtype}
\usepackage{inconsolata}

\usepackage{booktabs}
\usepackage{array}
\usepackage{colortbl}
\definecolor{groupbg}{HTML}{EEF1F4}
\definecolor{oursbg}{HTML}{EAF2F8}
\definecolor{lightrule}{HTML}{C9CED4}
\usepackage{enumitem}
\usepackage{graphicx}
\usepackage{xcolor}
\usepackage{pifont}
\usepackage[breakable,skins]{tcolorbox}
\usepackage{multirow}

\tcbset{promptstyle/.style={breakable, enhanced, colback=gray!4, colframe=gray!55, colbacktitle=gray!18, coltitle=black,
  boxrule=0.4pt, arc=1.5pt, left=3pt, right=3pt, top=2pt, bottom=2pt, toptitle=1pt, bottomtitle=1pt,
  fonttitle=\footnotesize\bfseries, before skip=6pt, after skip=8pt,
  before upper={\ttfamily\scriptsize\raggedright\setlength{\parindent}{0pt}\setlength{\parskip}{2pt}}}}
\newtcolorbox[auto counter]{promptbox}[2][]{promptstyle, title={Prompt~\thetcbcounter: #2}, #1}
\newtcolorbox[auto counter]{examplebox}[2][]{promptstyle, title={Example~\thetcbcounter: #2}, #1}
\newcommand{\pnote}[1]{{\normalfont\itshape\color{black!60}#1}}

\newcommand{\cmark}{\ding{51}}
\newcommand{\xmark}{\ding{55}}

\newcommand{\method}{3DCodeEdit}

\title{\method{}: Benchmarking Instruction-Guided 3D Program Editing}

\author{
  Binjie Wang\textsuperscript{1*} \quad Xinran Zhao\textsuperscript{2*} \quad Shoubin Yu\textsuperscript{3} \quad Xiang Li\textsuperscript{4} \quad Silong Yong\textsuperscript{2} \\
  \bfseries Chenyang Yang\textsuperscript{2} \quad Tianyu Chen\textsuperscript{5} \quad Ruofei Du\textsuperscript{6} \\
  \bfseries He He\textsuperscript{1} \quad Chen Zhao\textsuperscript{1} \quad Tongshuang Wu\textsuperscript{2} \\[6pt]
  {\normalfont\url{https://binjiew.github.io/3DCodeEdit}} \\[6pt]
  \normalfont\textsuperscript{1}New York University \quad \textsuperscript{2}Carnegie Mellon University \quad \textsuperscript{3}UNC, Chapel Hill \\
  \normalfont\textsuperscript{4}University of Illinois Urbana-Champaign \quad \textsuperscript{5}The University of Texas at Austin \quad \textsuperscript{6}Google \\[3pt]
  \normalfont\small\textsuperscript{*}Equal contribution
}

\begin{document}

\makeatletter\providecommand\captionof[1]{\def\@captype{#1}\caption}\makeatother
\twocolumn[{%
  \renewcommand\twocolumn[1][]{#1}%
  \maketitle
  \vspace{-1.2em}
  \begin{center}
    \includegraphics[width=\textwidth]{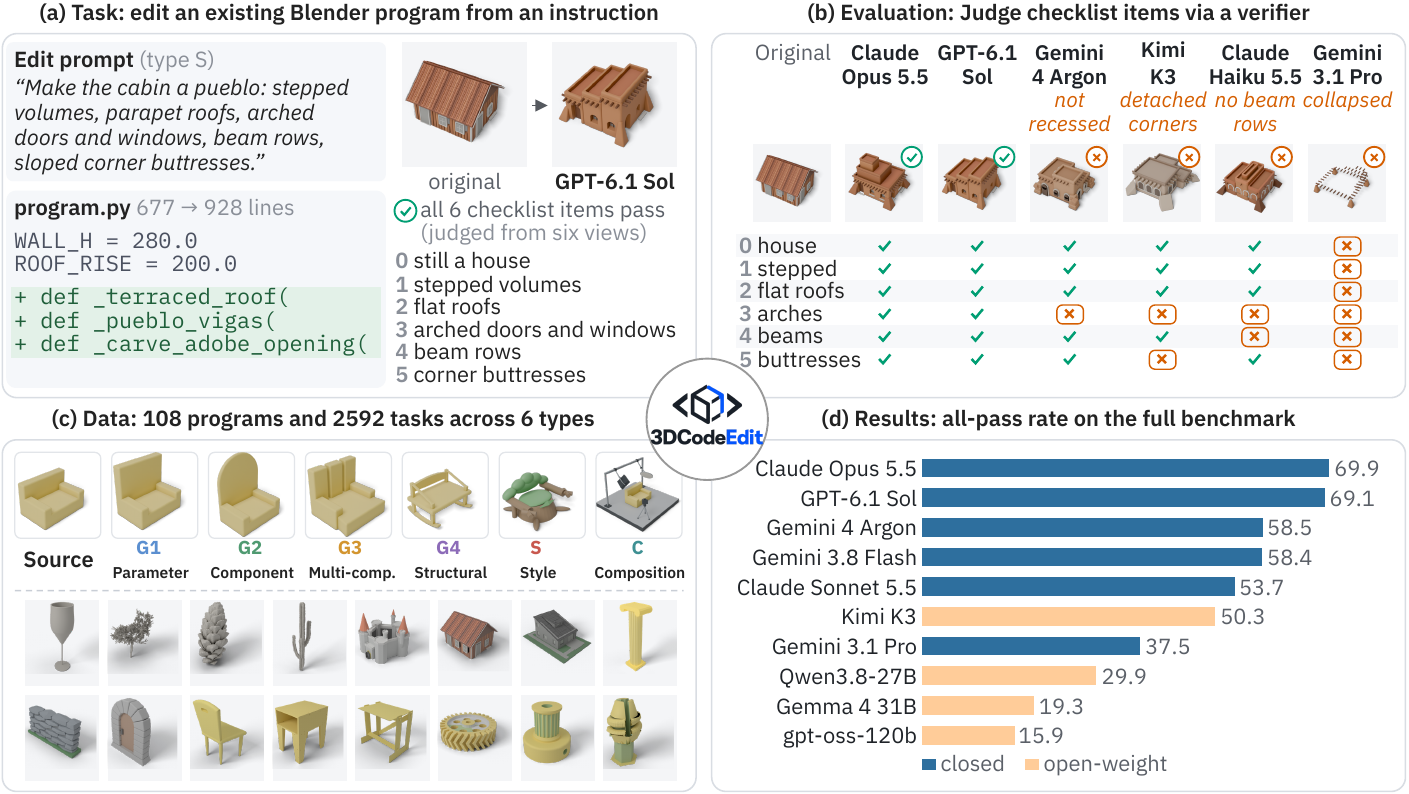}
    \captionof{figure}{We introduce \method{}, a benchmark for instruction-guided editing of 3D objects built in code. (a) Our task involves modifying an existing Blender program according to an edit instruction. (b) A verifier judges edited objects from rendered views against a checklist of requested changes and preservation requirements. (c) Six edit types range from parameter changes to structural edits, style redesign, and scene composition. The benchmark contains 2,592 editing tasks over 108 diverse source programs. (d) 3D program editing remains challenging: the best model achieves an all-pass rate of only 69.9\%.}
    \label{fig:teaser}
  \end{center}
  \vspace{0.8em}
}]

\begin{abstract}
\looseness=-1
Language models increasingly create 3D content through code, with practical workflows often relying on editing existing programs rather than regenerating them, for speed and precise control. Such edits require understanding what a program builds and foreseeing how a change reshapes it, yet prior benchmarks mostly evaluate generation or simple edits against a reference. In this paper, we introduce \method{}, a benchmark of 2,592 instruction-guided editing tasks over 108 Blender programs that construct objects entirely in code. 
Its edits range from parameter to structural changes, style redesign, and scene composition, and are judged from renders against human-reviewed checklists. Across 17 models, the average success rate is only 37\%, and even the strongest models complete fewer than 60\% of structural and style edits. 
As models become more capable, their failures shift to geometric foresight, where the code runs but builds the wrong shape. 
We therefore study whether \textbf{multi-turn feedback} can compensate for this missing foresight. Feedback from a separate reviewer model that describes how the renders differ from the request helps most, more than maximum reasoning effort or an agent harness, and can let a 27B model surpass the single-shot rate of a 2.8T-parameter model. 
\end{abstract}

\section{Introduction}
\label{sec:introduction}

\begin{table*}[t]
\centering
\renewcommand{\cmark}{\textcolor[HTML]{009E73}{\ding{51}}}%
\renewcommand{\xmark}{\textcolor[HTML]{D55E00}{\ding{55}}}%
\small
\setlength{\tabcolsep}{7pt}
\newcommand{\bchead}[1]{\begin{tabular}[c]{@{}c@{}}#1\end{tabular}}%
\begin{tabular}{lcccccc}
\toprule
 & Task & \multicolumn{3}{c}{Programs} & \multicolumn{2}{c}{Evaluation} \\
\cmidrule(lr){2-2}\cmidrule(lr){3-5}\cmidrule(lr){6-7}
Benchmark & Editing & \bchead{End-to-End\\Code} & \bchead{Complex\\Code} & \bchead{Multiple\\Sources} & \bchead{Difficulty\\Levels} & \bchead{Reference-Free\\Eval.} \\
\midrule
CADBench~\citep{du2024blenderllm}            & \xmark & \cmark & \xmark & \cmark & \xmark & \cmark \\
3DCodeBench~\citep{gao20263dcodebench}       & \xmark & \cmark & \cmark & \xmark & \xmark & \xmark \\
3DHarnessBench~\citep{liu2026harnessbench}   & \xmark & \cmark & \cmark & \xmark & \xmark & \xmark \\
Nova3D-Bench~\citep{noor2026nova3d}          & \xmark & \cmark & \cmark & \xmark & \cmark & \cmark \\
BlenderGym~\citep{gu2025blendergym}          & \cmark & \xmark & \xmark & \xmark & \xmark & \xmark \\
BlenderBench~\citep{yin2026viga}             & \cmark & \xmark & \xmark & \cmark & \cmark & \xmark \\
\midrule
\rowcolor{oursbg}\textbf{\method{} (ours)} & \cmark & \cmark & \cmark & \cmark & \cmark & \cmark \\
\bottomrule
\end{tabular}
\caption{\looseness=-1
Comparison of benchmarks for Blender code. \method{} combines instruction-guided editing, end-to-end object construction, complex programs from multiple sources, graded edit difficulty, and reference-free evaluation. In \textit{end-to-end code}, the program builds the entire object; \textit{complex code} indicates an average program length exceeding 200 lines (Appendix~\ref{app:code-length}); \textit{reference-free evaluation} judges edits against the instruction rather than a reference output.}
\label{tab:benchmark-comparison}
\end{table*}

\looseness=-1
Code is becoming a common way to represent 3D assets. Procedural generators define objects as programs~\citep{raistrick2023infinigen}, game engines define reusable assets as parametric classes~\citep{epic2026blueprintclass}, and frontier models now build objects by writing Blender programs~\citep{anthropic2026creativework,openai2026astraarchviz}. 
Once written, such a program is revised many times, as users refine the asset, adapt it into variants, or update it to new requirements. 
A user who has generated an armchair, for example, may later want a taller backrest, or want to turn it into a rocking chair (Figure~\ref{fig:teaser}c). Editing the program is faster than regenerating the chair, and it changes only the requested parts while keeping the rest intact.

Such an edit is demanding because the instruction describes how the object should look, while the model can only change the code. The model must work out from the code what shape it builds, foresee how a change will reshape it, and write renderable code that makes exactly this change. The task therefore requires multimodal understanding and foresight that link code to 3D geometry, on top of code editing. Even as code generation improves rapidly~\citep{chen2021codex,jimenez2024swebench,deng2026swebenchpro,merrill2026terminalbench}, frontier models still struggle with instructed code editing~\citep{cassano2024canitedit,chi2026editbench} and with spatial reasoning~\citep{fu2024blink,yang2025thinking,yang2026mmsi,liu2026spatialbabel}. For example, a model may break the structure of the existing program so that it no longer runs, or, failing to foresee the result of its change, write code that runs but builds the wrong shape, such as a new part floating in the air, disconnected from the object~\citep{gao20263dcodebench}.

A benchmark for this kind of editing should meet three requirements. First, the program should build the entire object in code, exposing its parameters and parts to the edit. An object stored in a pre-built scene file or as raw mesh data leaves nothing meaningful to edit. Ideally, the programs are also larger, more complex, and diverse, so that the shape cannot be read off at a glance. Second, the edits should range from local changes to structural and open-ended ones, which capture more edit types in practice and also require foreseeing more of the geometry. Third, each edit should be judged from renders on whether the requested change is made and the rest preserved, without relying on a reference output. An instruction rarely fixes a single outcome. 
Even ``make the backrest taller'' can be met at many heights, and open-ended edits, common in real-world editing, admit far more. However, existing benchmarks either evaluate generation from scratch~\citep{du2024blenderllm,gao20263dcodebench,liu2026harnessbench,noor2026nova3d} or edit objects in pre-built scenes against a reference output~\citep{gu2025blendergym,yin2026viga} (Table~\ref{tab:benchmark-comparison}).

\begin{figure*}[t]
    \centering
    \includegraphics[width=\textwidth]{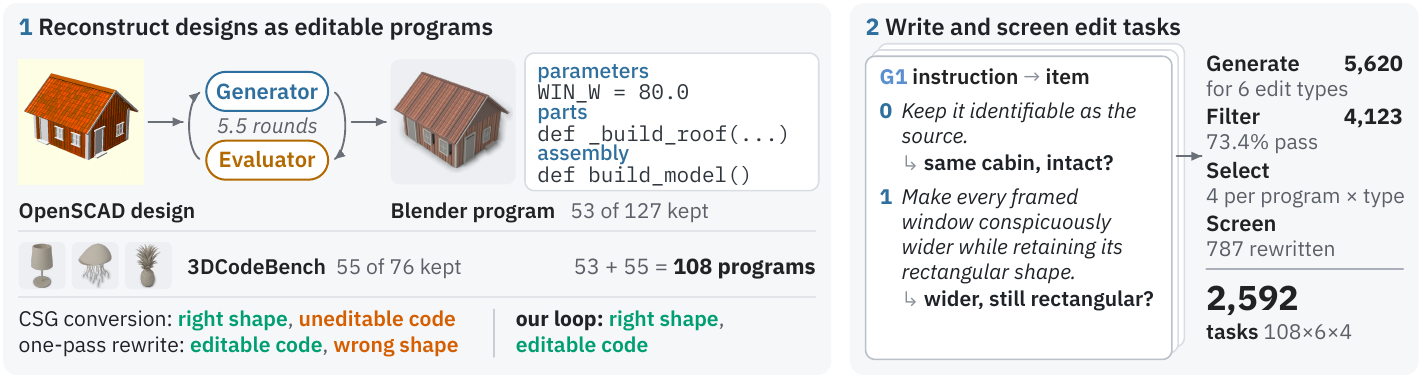}
    \caption{\looseness=-1
    Construction of \method{}. (1) A generator and an evaluator iteratively reconstruct OpenSCAD designs as editable Blender programs, preserving geometry while exposing parameters, parts, and assembly logic. We combine 53 reconstructed programs with 55 from 3DCodeBench. (2) For each program and edit type, we generate edit prompts paired with checklists of requested changes and preservation requirements. After filtering, selection, and ambiguity screening under human supervision, we obtain four tasks per program and edit type, totaling 2,592 tasks.}
    \label{fig:overview}
\end{figure*}

To fill this gap, we introduce \method{}, a benchmark of 2,592 instruction-guided editing tasks over 108 Blender programs (Figures~\ref{fig:teaser} and~\ref{fig:overview}). Each program builds its object entirely in code. Besides procedurally generated objects, the programs include designs of buildings, furniture, and mechanical parts shared by designers online, which we reconstruct as editable Blender programs through an agent loop grounded in renders. Building on these programs, we design six edit types: four geometric types graded by scope, from a single parameter to the structure of the whole object, and two open-ended types, style redesign and scene composition. The editor sees only the program and the instruction. An edit succeeds only if its code runs and renders, and it passes a human-reviewed checklist, with one item per requested change and one for preserving the rest, as judged by a multimodal verifier that compares renders of the original and edited objects.

We evaluate 17 models and find that editing remains hard, and gets harder with scope. The average success rate is only 37\%, falling from 82.9\% on single-parameter edits to 24.6\% on structural edits, and even strong models, Claude Opus 5.5 and GPT-6.1-sol, complete fewer than 60\% of structural and style edits. Moreover, as models get stronger, their failures shift from misreading the instruction or the program, or writing code that does not run, to geometric foresight, where the code runs but builds the wrong shape. Since the editor cannot foresee its result, we further study multi-turn editing, in which a reviewer inspects the renders and returns feedback, as a user would. We investigate which kinds of feedback compensate for the missing foresight and who should provide them. Seeing its own renders helps the editor, but models rarely revise their own edits. Feedback from a separate reviewer helps far more, especially when it describes how the renders differ from the request, and beats both maximum reasoning effort and an agent harness. The reviewer needs to judge well rather than edit well, and with a stronger reviewer, even Qwen3.8-27B surpasses the single-shot rate of the 2.8T-parameter Kimi K3 (50.3\% vs.\ 46.8\%). Our contributions are as follows:
\begin{itemize}[labelwidth=*,leftmargin=1.3em,align=left, topsep=0pt, nosep]
    \item We present \method{}, a novel benchmark that captures edits to how an object is built in code, graded from a single parameter to the whole structure, as well as open-ended style redesign and scene composition, which previous benchmarks do not test.
    \item We design a reference-free evaluation that makes open-ended edits measurable, judging each edit from renders against an instance-specific checklist of the requested changes and the preservation of the rest, with 90.4\% agreement with human labels.
    \item We evaluate 17 models and track 44 over time, revealing that as models improve, their failures shift from comprehension and coding to geometric foresight.
    \item We show that feedback from a separate reviewer, especially a description of how the renders differ from the request, compensates for this missing foresight far better than self-checking, more reasoning, or an agent harness, even when the reviewer edits worse than the editor.
\end{itemize}

\section{Related Work}
\label{sec:related-work}

\looseness=-1
\paragraph{Generating and editing 3D with code.}
Neural 3D generators now produce high-fidelity assets~\citep{poole2022dreamfusion,xiang2024trellis,zhao2025hunyuan3d20}, but they represent an asset without explicit parts or parameters, so precise and controllable edits remain difficult~\citep{zhang2025shapecraft,yang2025omnipart,sella2026proxe,xia2025scalable}. Code instead exposes parameters and construction logic. Language models first generated 3D programs by calling procedural generators such as Infinigen~\citep{raistrick2023infinigen,sun2025threedgpt} or simplified APIs~\citep{yamada2025l3go}, and now write Blender programs directly, building complex, part-structured objects and scenes~\citep{hu2024scenecraft,du2024blenderllm,lu2025ll3m,dai2025meshcoder,noor2026nova3d,yin2026viga}. Editing has received less attention. BlenderAlchemy searches over edits to materials, geometry, and lighting~\citep{huang2024blenderalchemy}, and LL3M, MeshCoder, and Nova3D demonstrate edits through code changes~\citep{lu2025ll3m,dai2025meshcoder,noor2026nova3d}, but these works evaluate editing only qualitatively or on small, narrow sets.

\looseness=-1
\paragraph{Benchmarks for 3D code.}
Most benchmarks for Blender code evaluate generation from scratch, from text, images, target geometry, or videos~\citep{du2024blenderllm,gao20263dcodebench,liu2026harnessbench,noor2026nova3d,tang2026bvb}. Only two benchmarks evaluate editing, both at a small scale. BlenderGym defines each of its 245 instances by a start and a goal scene and scores an edit by its distance to the goal~\citep{gu2025blendergym}, and BlenderBench specifies 27 tasks by target images~\citep{yin2026viga}. Both run short scripts on pre-built scene files, and their edits tune placement, materials, lighting, the camera, shape keys, or the parameters of an existing generator, without restructuring how an object is built. Their reference outputs also penalize valid edits that differ from the target, which rules out open-ended edits. \method{} instead edits programs that build the whole object, at scopes up to its overall structure and including open-ended edits, which it judges with instance-specific checklists (Table~\ref{tab:benchmark-comparison}).

\section{The \method{} Benchmark}
\label{sec:benchmark}

\looseness=-1
\method{} pairs Blender programs that build entire objects with instructed edits of graded scope, and judges each edit from renders against a checklist (Figure~\ref{fig:overview}). After defining the task (\S\ref{sec:task}), we describe the programs (\S\ref{sec:construction}), the edits (\S\ref{sec:levels}), and the checklists that judge them (\S\ref{sec:checklists}).

\subsection{Task Definition}
\label{sec:task}
\label{sec:protocol}
\looseness=-1
Given a Blender program that builds an object and an edit prompt, the editor returns the edited program. The edit succeeds if the edited program runs and renders, and its renders satisfy every item of a checklist paired with the prompt.

\looseness=-1
Formally, a task is a tuple $(P, q, L)$. $P$ is a Blender Python program, and $R(P)$ denotes its renders from six fixed views. The edit prompt $q=(q_0,q_1,\dots,q_k)$ consists of a preservation instruction $q_0$, which asks to keep the object's identity and the parts outside the requested edits, and edit instructions $q_1,\dots,q_k$. The checklist $L=(\ell_0,\dots,\ell_k)$ pairs each instruction $q_i$ with a binary item $\ell_i$. The editor receives only $(P,q)$ and returns an edited program $P'$. A verifier receives $R(P)$, $R(P')$, $q$, and one item $\ell_i$, but neither program, and returns $v_i\in\{0,1\}$. The edit succeeds if
\begin{equation}
  y = \mathbf{1}[R(P') \neq \emptyset] \cdot \textstyle\prod_{i \notin A} v_i ,
  \label{eq:success}
\end{equation}
\looseness=-1
where $R(P')=\emptyset$ if $P'$ fails to execute or render, and $A$ is a small set of items excluded as ambiguous (\S\ref{sec:checklists}). We report the \emph{all-pass rate}, the mean of $y$ over tasks. Both objects are rendered with the same cameras, so the verifier can judge comparative items such as ``taller than before'' and detect unintended changes (Appendix~\ref{app:protocol}).

\subsection{Programs That Build Entire Objects}
\label{sec:construction}
\looseness=-1
Our 108 programs cover six object categories (Figure~\ref{fig:data-stats}a), with 53 reconstructed from OpenSCAD\footnote{\url{https://openscad.org}} models and 55 from 3DCodeBench~\citep{gao20263dcodebench}. From 3DCodeBench, we keep 55 of 76 executed candidates that have sound geometry and offer visible structure to edit. Since these programs include no buildings or mechanical parts, we reconstruct the other 53 from parametric OpenSCAD models of architecture, furniture, and manufacturing objects shared by designers online.

\looseness=-1
Converting these models directly loses either the parameters or the shape. Translating the constructive solid geometry keeps the shape but flattens modules and parameters into raw vertices and Boolean operations, whereas asking a model to rewrite the design in one pass keeps the structure but loses fidelity on complex objects (Appendix~\ref{app:sources}). We instead reconstruct each model in a loop grounded in renders (Figure~\ref{fig:overview}). A generator agent writes and renders a Blender program, and an independent evaluator agent compares its renders with those of the source and either accepts it or returns feedback for the next round. Mirroring how designers write parametric models, the generator separates low-level geometry helpers from a high-level, parameterized assembly of the parts, so that edits of growing scope touch a parameter, a part, or the assembly itself. Reconstructions take 5.5 rounds on average and run without human intervention (example trajectory in Appendix Figure~\ref{fig:recon-trajectory}). From 127 reconstructions, we keep 53 after removing simple objects, duplicate object types, and models whose license does not clearly permit redistribution. The resulting programs (examples in Appendix Figure~\ref{fig:source-gallery}) have a median of 440 lines, an order of magnitude longer than the BlendNet programs used by CADBench (Appendix~\ref{app:code-length}).

\subsection{Edits of Graded Scope}
\label{sec:levels}
\begin{figure*}[t]
    \centering
    \includegraphics[width=\textwidth]{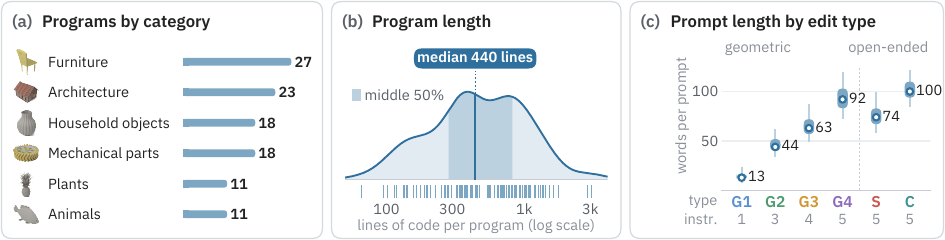}
    \caption{Statistics of \method{}. (a) The 108 source programs by object category. (b) Program length in lines of code (log scale; shaded: middle 50\%; ticks: individual programs). (c) Words per edit prompt by edit type (G1--G4, S, C; defined in Table~\ref{tab:taxonomy}; dot: median; bar: interquartile range; whisker: 5th--95th percentile), with the number of edit instructions below.}
    \label{fig:data-stats}
\end{figure*}

\looseness=-1
The scope of an edit determines how much of the existing geometry the model must work out and foresee. We therefore grade edits by scope, following the SOLO taxonomy of learning outcomes~\citep{biggs1982solo}, and add open-ended edits that go beyond the existing structure, as users often request in practice (Table~\ref{tab:taxonomy}). The four geometric edit types grow in scope. G1 changes one attribute of a part, G2 and G3 rebuild one or two parts, and G4 reorganizes how parts relate to one another. Style redesign (S) gives the whole object a different style, and scene composition (C) builds a functional scene around it. Each program receives four prompts per edit type, giving 2,592 tasks (Figure~\ref{fig:data-stats}c).

\begin{table}[t]
\centering
\small
\setlength{\tabcolsep}{4pt}
\begin{tabular}{cl>{\raggedright\arraybackslash}p{0.50\columnwidth}}
\toprule
Type & Name & Operational definition \\
\midrule
\rowcolor{groupbg}\multicolumn{3}{l}{\textit{Geometric edits, graded by scope}} \\
G1 & Parameter & Change one visible attribute (size, proportion, spacing, position, or orientation) of one part or part family. \\
G2 & Component & Reconstruct the form of one part, together with a size change and one further visible change to it. \\
G3 & Multi-component & Reconstruct and resize two separate parts independently. \\
G4 & Structural & Reorganize how three or more regions meet, span, or support one another, changing the overall layout. \\
\rowcolor{groupbg}\multicolumn{3}{l}{\textit{Open-ended edits}} \\
S & Style & Redesign the object in a different style by reshaping its main body and adding four large, mostly geometric cues. \\
C & Composition & Place the object in a functional scene with a support surface, three working props, and one large context structure. \\
\bottomrule
\end{tabular}
\caption{\looseness=-1
Edit types in \method{} and their operational definitions.}

\label{tab:taxonomy}
\end{table}

\subsection{Checklists Judged from Renders}
\label{sec:checklists}
\label{sec:quality}
\label{sec:stats}
\looseness=-1
Since an instruction admits many valid results, we score no edit against a reference output. Instead, every instruction $q_i$ is paired with a checklist item $\ell_i$ that verifies it, so each item must be decidable from the renders and admit a single reading. Appendix~\ref{app:case-gallery} shows eight tasks with edits from six models.

\paragraph{Writing and filtering prompts.}
For each program and edit type, GPT-5.6-sol reads the program, its description, and four renders, and writes eight candidate prompts, pairing each instruction with a checklist item. Each task also receives a preservation instruction $q_0$, whose item $\ell_0$ checks that the object stays identifiable outside the edit. We then filter the candidates in three steps. (1) GPT-5.6-sol rejects candidates whose outcome would not be visible or that violate the edit-type definition. (2) Human annotators reject candidates that are ambiguous, not visually verifiable, or unreasonable. Together, these two steps admit 73.4\% of 5,620 candidates, and we keep four per program and edit type, giving 2,592 prompts with 12,528 checklist items. (3) Claude Opus 5.5 screens every item for multiple readings, such as whether ``about 50\% larger'' refers to diameter or area, and flagged instructions are rewritten under human supervision. This revises 787 prompts, and the 51 items (0.4\%) that remain ambiguous are excluded from scoring as the set $A$. Details are in Appendices~\ref{app:instructions} and~\ref{app:ambiguity}.

\begin{table*}[t]
\centering
\small
\setlength{\tabcolsep}{5.5pt}
\begin{tabular}{
  ll
  !{\color{lightrule}\vrule}
  l@{\,}rrrr
  !{\color{lightrule}\vrule}
  l@{\,}rr
  !{\color{lightrule}\vrule}
  rr
}
\toprule
& Model
& {\scriptsize\textcolor{gray}{Geometric}}
& G1 & G2 & G3 & G4
& {\scriptsize\textcolor{gray}{Open-ended}}
& S & C & All & Exec. \\
\midrule
\multirow{9}{*}{\textit{Proprietary}}
& Claude Opus 5.5 & & \textbf{96.5} & \textbf{76.9} & 64.4 & 58.6 & & 52.5 & \textbf{70.8} & \textbf{69.9} & \textbf{99.3} \\
& GPT-6.1-sol & & 95.6 & 71.3 & \textbf{65.3} & \textbf{59.3} & & \textbf{57.6} & 65.7 & 69.1 & 96.1 \\
& Gemini 4 Argon & & 94.7 & 67.4 & 54.2 & 41.0 & & 41.0 & 53.0 & 58.5 & 95.4 \\
& Gemini 3.8 Flash & & 94.7 & 68.8 & 62.7 & 53.7 & & 34.5 & 36.3 & 58.4 & 92.0 \\
& Claude Sonnet 5.5 & & 91.4 & 66.4 & 51.2 & 41.9 & & 25.9 & 45.4 & 53.7 & 98.8 \\
& Claude Haiku 5.5 & & 90.3 & 54.4 & 42.8 & 31.5 & & 16.0 & 32.9 & 44.6 & 94.3 \\
& GPT-6-luna & & 87.0 & 51.4 & 36.6 & 29.4 & & 18.8 & 26.9 & 41.7 & 92.4 \\
& Gemini 3.1 Pro & & 88.7 & 50.9 & 41.0 & 28.0 & & 7.9 & 8.3 & 37.5 & 96.4 \\
& Claude Haiku 4.5 & & 81.5 & 28.9 & 9.3 & 4.4 & & 0.0 & 0.0 & 20.7 & 91.6 \\
\midrule
\multirow{8}{*}{\textit{Open-weight}}
& Kimi K3 & & 88.2 & 57.9 & 47.5 & 38.9 & & 27.3 & 42.4 & 50.3 & 96.5 \\
& Qwen3.8-27B & & 84.3 & 44.2 & 28.5 & 17.1 & & 3.5 & 1.6 & 29.9 & 89.3 \\
& Qwen3.6-27B & & 80.8 & 29.6 & 16.0 & 6.0 & & 0.2 & 0.0 & 22.1 & 81.8 \\
& Gemma 4 31B & & 77.5 & 21.5 & 11.8 & 4.9 & & 0.2 & 0.0 & 19.3 & 86.2 \\
& Qwen3.6-35B-A3B & & 70.6 & 18.8 & 5.3 & 1.2 & & 0.0 & 0.0 & 16.0 & 72.7 \\
& gpt-oss-120b & & 71.5 & 18.1 & 4.4 & 1.6 & & 0.0 & 0.0 & 15.9 & 77.3 \\
& Qwen3-VL-235B-A22B & & 61.8 & 9.0 & 2.3 & 0.5 & & 0.0 & 0.0 & 12.3 & 68.4 \\
& Qwen3-Coder-Next & & 54.2 & 4.9 & 1.4 & 0.2 & & 0.0 & 0.0 & 10.1 & 51.1 \\
\midrule
\multicolumn{2}{l!{\color{lightrule}\vrule}}{Average}
& & 82.9 & 43.5 & 32.0 & 24.6 & & 16.8 & 22.5 & 37.1 & 87.0 \\
\bottomrule
\end{tabular}
\caption{\looseness=-1
Single-shot editing performance (\%) on \method{}. All-pass rate measures the percentage of edits whose programs execute, render, and satisfy every checklist item. G1–G4 denote geometric edits of increasing scope; S and C denote style and composition edits. \textit{Exec.} reports execution and rendering success. Average is the mean over the 17 models. Five Qwen3.5 models are reported in Table~\ref{tab:main-results-full}.}
\label{tab:main-results}
\end{table*}

\paragraph{Verifier.}
Among 16 multimodal models compared against human judgments, we choose Gemini 3.8 Flash as the verifier to balance accuracy and cost (Appendix~\ref{app:verifier}). On 210 items that a human annotator labels from the same renders, it agrees with the human label on 90.4\% of items (Cohen's $\kappa=0.80$). Judging from the programs instead lowers agreement for every verifier we test, because code that reads as correct can still build the wrong shape, such as a part tilted the wrong way or left floating, which is hard to see in the code but obvious in the renders (Appendix~\ref{app:verifier}).

\section{How Well Do Models Edit 3D Programs?}
\label{sec:exp-main}
\looseness=-1
\paragraph{Setup.}
We evaluate 17 models, 9 proprietary and 8 open-weight, in a single-shot setting: each model receives the task once and returns the complete edited program. We report the all-pass rate (\S\ref{sec:task}): the percentage of tasks whose edit renders and passes every checklist item under the verifier of \S\ref{sec:quality}. Model references and decoding settings are in Appendix~\ref{app:protocol}. Analyses that need many runs use an \emph{analysis subset} of 312 tasks from 13 randomly chosen source programs (\S\ref{sec:exp-timeline}, \S\ref{sec:exp-errors}, and \S\ref{sec:exp-multiturn}).

\subsection{Main Results}
\label{sec:exp-results}
\looseness=-1
Table~\ref{tab:main-results} reveals three patterns.
\textbf{First, a steep scope gap.} The average all-pass rate is 37.1\%, and even the strongest editors, Claude Opus 5.5 and GPT-6.1-sol, reach only about 70\% overall and fewer than 60\% on structural (G4) and style (S) edits. The average rate drops most from parameter edits (G1, 82.9\%) to component edits (G2, 43.5\%) and falls to 24.6\% on structural edits (G4).
\textbf{Second, open-ended edits distinguish strong models.} All six types order the 17 models almost the same way (mean Spearman $\rho=0.97$), but the gaps between models are not the same. Claude Opus 5.5, GPT-6.1-sol, and Gemini 3.8 Flash are within about 8 points of each other on G2--G4, yet 23 points apart on S and 35 points apart on C.
\textbf{Third, the two best models solve different tasks.} Claude Opus 5.5 and GPT-6.1-sol each solve about 70\% of the tasks, but 82.6\% are solved by at least one of them. Their successes overlap least on S and C (task-level correlation 0.23 and 0.18, against about 0.4 on G2--G4; Appendix Table~\ref{tab:complementarity}).

\begin{figure*}[t]
    \centering
    \includegraphics[width=\textwidth]{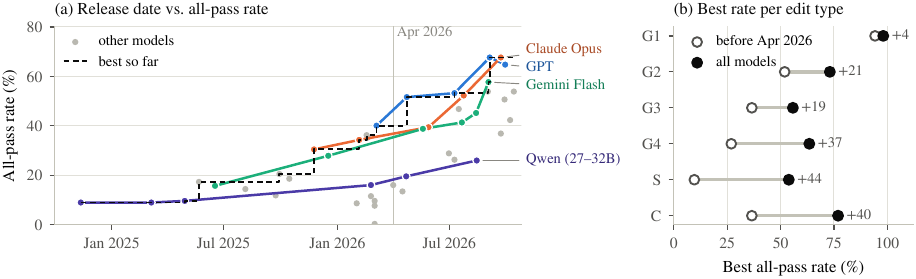}
    \caption{\looseness=-1
    Progress in 3D program editing
    on the analysis subset (312 tasks). (a) All-pass rate of 44 models; lines join releases within four families, and the dashed line is the best rate so far. (b) Best rate per edit type among models released before April 2026 and among all models.}
    \label{fig:timeline}
\end{figure*}

\subsection{Editing Ability over Time}
\label{sec:exp-timeline}
\looseness=-1
To measure how fast this ability improves, we evaluate 44 models released since November 2024 on the analysis subset (Figure~\ref{fig:timeline}; models listed in Appendix Table~\ref{tab:timeline-models}).
\textbf{First, progress has accelerated.} The best all-pass rate took 16 months to rise from 9.0\% to 40.1\% (November 2024 to March 2026) and only six more months to reach 67.6\%. The same holds within families: Claude Opus gained 9 points in the six months up to Opus 4.8 and 28 points in the four months after it.
\textbf{Second, mid-sized open-weight models fall behind.} The 27--32B Qwen models trail the best model available at their release by 20 points (Qwen3.5, February 2026) and 27 points (Qwen3.8, August 2026). \textbf{Third, recent gains appear on open-ended edits.} Earlier models could already handle simple edits but rarely open-ended ones. Claude Opus 4.5 (November 2025), for example, passes 88.5\% of parameter edits but under 4\% of style and composition edits. Ten months later, Claude Opus 5.5 passes 51.9\% of style and 76.9\% of composition edits, matching or exceeding its 51.9\% on structural edits, which were once easier for models (Appendix Table~\ref{tab:timeline-models}). Structural edits, in contrast, must foresee how several tightly coupled parts interact, which remains a bottleneck (\S\ref{sec:exp-errors}).

\begin{table}[!t]
\centering
\small
\setlength{\tabcolsep}{3.5pt}
\begin{tabular}{@{}lrrrrrr@{}}
\toprule
 & & \multicolumn{5}{c}{Share of failures} \\
\cmidrule(l){3-7}
All-pass band & $n$ & I & P & K & G & O \\
\midrule
$<$20\% & 19 & 11\% & 34\% & 37\% & 18\% & 0\% \\
20--40\% & 19 & 11\% & 29\% & 23\% & 37\% & 0\% \\
40--55\% & 23 & 10\% & 24\% & 22\% & 42\% & 2\% \\
$\geq$55\% & 15 & 7\% & 27\% & 23\% & 39\% & 4\% \\
\bottomrule
\end{tabular}
\caption{\looseness=-1
Distribution of failure stages by editing performance. Each failure is assigned to the first stage at which the edit diverges from the instruction. \textbf{I}: instruction comprehension; \textbf{P}: program comprehension; \textbf{K}: code execution; \textbf{G}: geometric foresight; \textbf{O}: visibility in renders. As performance improves, failures shift from comprehension and coding toward geometric foresight.}
\label{tab:editor-errors}
\end{table}

\subsection{Why Do Models Fail?}
\label{sec:exp-errors}
\looseness=-1
We code every failure of all 76 configurations on the analysis subset (the 44 models of Figure~\ref{fig:timeline} and 32 further effort levels of nine models) with a two-level taxonomy: the stage at which the edit first diverges from the instruction, and a mechanism within that stage (Appendix Table~\ref{tab:taxonomy-full}; details in Appendix~\ref{app:errors}). Edits that fail to run or render are classified from their error logs, and the others by Gemini 3.8 Flash from the instruction, both programs, and both sets of renders (Cohen's $\kappa=0.71$ against human labels on 150 failures; Appendix~\ref{app:errors}). Table~\ref{tab:editor-errors} summarizes the result.

\paragraph{As models improve, failures shift from comprehension and coding to geometric foresight.}
Most failures of weak models happen before the shape is built, as the model misreads the instruction or the program, or writes code that does not run (82\% of failures in the weakest band; Table~\ref{tab:editor-errors}). Strong models make these mistakes less often, and their failures move to geometric foresight. Their code runs and changes the intended parts, but builds the wrong shape. Avoiding them requires 3D understanding and spatial reasoning rather than coding skill. Therefore, such failures are hard to detect by running or reading the code, and show up only in the renders. The shift follows capability. From the weakest band to the two strongest, comprehension and coding fall to under 60\% of failures, while geometric foresight doubles from 18\% to about 40\% and becomes the most common stage (Table~\ref{tab:editor-errors}).

\looseness=-1
\paragraph{Geometric foresight fails most where parts meet.}
Looking further into geometric foresight (G) across all models, we find that the most common mechanism is an unforeseen interaction between parts (G-INTERACT, 9.2\% of all failures), where a new or changed part collides with or is swallowed by the existing geometry, or leaves an unintended gap. Next come generator parameters that reshape the object in unintended ways (G-PARAM, 7.3\%) and parts that float or detach (G-SUPPORT, 5.3\%). Edits of the wrong size, too small or too large, are less than half as frequent as interactions (G-TIMID and G-OVER, 4.3\% together; Appendix Table~\ref{tab:taxonomy-full}). Models thus usually make the requested change but misjudge how it fits with the rest of the object.

\subsection{Multi-Turn Editing: Supplying What the Editor Cannot Foresee}
\label{sec:exp-multiturn}

\looseness=-1
\S\ref{sec:exp-errors} suggests that strong models most often fail to foresee what their code builds. If foresight is the main bottleneck, supplying the result should help more than additional reasoning or attempts without it. We test this through multi-turn editing, which mirrors how a user inspects the result and asks for corrections. After each edit, the editor may receive feedback and revise its program, for up to 8 rounds on the analysis subset (details in Appendix~\ref{app:multiturn}). Our settings compensate for the missing foresight in different ways, each supplying a different amount of information about the result (Table~\ref{tab:mt-main}):
\begin{enumerate}[leftmargin=*,itemsep=0pt,parsep=0pt,topsep=2pt,partopsep=0pt]
\looseness=-1
    \item \emph{Self-check (code)}: no new information. The editor rereads its own code and decides whether to revise, which controls for the extra rounds alone.
\looseness=-1
    \item \emph{Self-check (renders)}: the result itself. The editor sees its renders but judges them on its own.
\looseness=-1
    \item \emph{Reviewer, P/F}: which requirements the result misses. A separate reviewer, GPT-6-luna, writes its own checklist from the instruction and the original renders, inspects the edited renders as a user would, and reports only which items pass.
\looseness=-1
    \item \emph{Reviewer, full}: how the result misses them. The reviewer also describes how the renders differ from what each item requires.
\looseness=-1
    \item \emph{Self-review}: the same full feedback, from the editor's own model in a separate conversation.
\end{enumerate}

\begin{table}[t]
\centering
\footnotesize
\setlength{\tabcolsep}{2pt}
\resizebox{\columnwidth}{!}{%
\begin{tabular}{@{}lrrrrr@{\hspace{8pt}}r@{}}
\toprule
 & & \multicolumn{2}{c}{Self-check} & \multicolumn{2}{c}{Indep.\ rev.} & \\
\cmidrule(lr){3-4}\cmidrule(lr){5-6}
Editor & T0 & Code & Render & P/F & Full & Self-rev. \\
\midrule
GPT-6.1-sol & 64.7 & 71.5 & 76.0 & 75.6 & 85.6 & 83.0 \\
Claude Opus 5.5 & 67.6 & 68.3 & 74.0 & 81.7 & 83.0 & 83.3 \\
Claude Sonnet 5.5 & 50.6 & 55.1 & 55.4 & 67.3 & 71.2 & 69.9 \\
GPT-6-luna & 36.9 & 42.6 & 49.7 & 59.0 & 65.7 & 61.5 \\
\arrayrulecolor{lightrule}\midrule\arrayrulecolor{black}
Qwen3.8-27B & 26.0 & 31.4 & 33.0 & 43.6 & 50.3 & 37.2 \\
Qwen3-VL-235B & 11.9 & 10.9 & 12.5 & 14.4 & 16.7 & 13.1 \\
gpt-oss-120b & 14.4 & 15.4 & -- & 17.9 & 22.8 & 15.1 \\
Qwen3-Coder-Next & 8.7 & 10.6 & -- & 13.1 & 13.5 & 10.6 \\
\bottomrule
\end{tabular}}
\caption{All-pass rate (\%) of multi-turn editing on the analysis subset (\S\ref{sec:exp-multiturn}). \textit{T0}: single-shot edit. \textit{Indep.\ rev.}: independent reviewer (GPT-6-luna, or GPT-6.1-sol for GPT-6-luna). \textit{Self-rev.}: self-review. --: render self-check unavailable for text-only editors.}
\label{tab:mt-main}
\end{table}

\looseness=-1
\paragraph{Seeing the result helps.}
When the editor checks its own work, seeing its renders beats rereading the code for every editor that can see images, by up to 7.1 points for GPT-6-luna (Table~\ref{tab:mt-main}). This gain comes from seeing the result itself, consistent with foresight being a bottleneck.

\looseness=-1
\paragraph{A reviewer's description of the renders helps most.}
However, strong models rarely revise after checking their own renders, in at most 0.6 rounds on average, against 2.7 or more with a reviewer (Appendix Table~\ref{tab:mt-rounds}). The more a reviewer tells the editor, the better it edits. Reporting only which items pass already beats self-check with renders for most editors, and describing how the renders differ from each requirement adds up to 10 more points, as for GPT-6.1-sol (75.6\% to 85.6\%). A description states exactly what the editor fails to foresee, whereas renders alone leave it to find the difference itself. For text-only editors, a reviewer's feedback is even the only way to learn what their code builds, and full feedback raises gpt-oss-120b from 14.4\% to 22.8\%.

\looseness=-1
\paragraph{Feedback beats more reasoning and an agent harness.}
Feedback also helps more than additional reasoning. Full feedback lifts GPT-6.1-sol from 64.7\% to 85.6\%, whereas its best reasoning effort reaches only 74.7\% in a single shot (Appendix Table~\ref{tab:effort-sweep}). Similarly, an agent harness that renders its own result and decides when to stop falls short, unless a reviewer provides feedback and decides when the edit is done (Appendix~\ref{app:codex}). The gains thus come from a review of the result rather than from additional computation or the renders alone, consistent with foresight, rather than writing the code, being the main bottleneck.

\looseness=-1
\paragraph{A reviewer needs to judge well, not necessarily edit well.}
Who, then, should provide the feedback? For strong editors, a weaker model is enough. Although GPT-6-luna edits worse, it reviews the three strongest editors about as well as self-review does, so judging an edit is easier than producing it. Weak editors, however, cannot judge their own edits. With GPT-6-luna instead of self-review, Qwen3.8-27B rises from 37.2\% to 50.3\%, surpassing the single-shot rate of Kimi K3 (46.8\%).

\section{Conclusion}
\label{sec:conclusion}
\looseness=-1
We introduced \method{}, a benchmark for editing Blender programs that build entire objects. Its edits range from geometric changes graded by scope to open-ended style redesign and scene composition, and each edit is judged from renders against a checklist. Editing remains hard, especially for structural and open-ended edits, and as models improve, their failures shift from comprehension and coding to geometric foresight. Supplying the result through multi-turn feedback closes much of this gap. A reviewer's description of the renders helps most, more than self-checking, maximum reasoning effort, or an agent harness. These results point to geometric foresight as an important capability to improve, and \method{} provides a testbed for measuring it.

\section*{Limitations}
\label{sec:limitations}

\looseness=-1
\method{} covers Blender programs that build single objects. Other 3D tools, such as CAD scripts, game-engine blueprints, and shader code, as well as larger scenes, animation, and articulated objects, are natural extensions of the same construction and evaluation pipeline. Our multi-turn study examines feedback in natural language, and other forms of feedback, such as sketches or reference images, remain to be explored. Finally, the checklists target the requested changes and the preservation of the object, so they do not capture every aspect of quality that a 3D artist would assess, such as aesthetic appeal. We release our code and data so that future work can extend these dimensions.

\section*{Ethical Considerations}
\label{sec:ethics}

\looseness=-1
All source programs are derived from publicly available models whose licenses permit redistribution, and we release the benchmark under the original licenses with attribution. Edit prompts, checklists, and reconstructions were produced with language models. Human annotators reviewed the prompts and checklists. The benchmark consists of everyday objects such as furniture, buildings, and mechanical parts, and we do not foresee risks beyond those of general-purpose code generation. We used AI assistants for coding and for editing the writing of this paper, and the authors take full responsibility for all content.

\section*{Acknowledgments}
\label{sec:acknowledgments}

We thank Chris Donahue, Azade Nova, Lijun Yu, Hongming Zhang, Yu She, Alex Wang, Yu Quan Chong, William Zhang, and Yuhao Peng for constructive discussions. This work was supported by grants from the Gemini Academic Program Award and the Amazon Research Award. Xinran Zhao's research is supported by the Amazon AI PhD Fellowship. This work was supported in part through the NYU IT High Performance Computing resources, services, and staff expertise.

\bibliography{references}

\clearpage
\appendix
\section{Benchmark Construction Details}
\label{app:construction}

\subsection{Source Programs}
\label{app:sources}
\looseness=-1
\paragraph{3DCodeBench.}
We execute and render 76 candidate programs from 3DCodeBench and keep the 55 that run, have sound geometry, do not duplicate another selected object, and offer visible structure to edit. The retained programs are used without modification.

\begin{figure}[t]
    \centering
    
    \begin{minipage}[b]{0.24\linewidth}\centering\includegraphics[width=\linewidth]{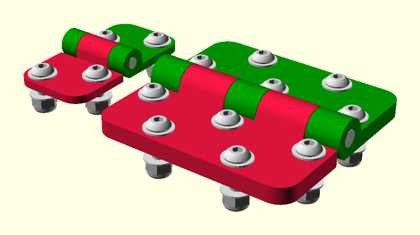}\\[1pt]{\scriptsize OpenSCAD}\end{minipage}\hfill
    \begin{minipage}[b]{0.24\linewidth}\centering\includegraphics[width=\linewidth]{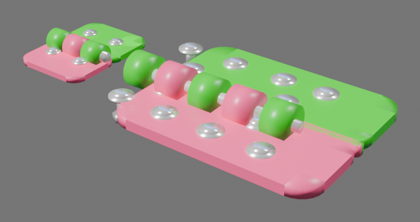}\\[1pt]{\scriptsize Round 0}\end{minipage}\hfill
    \begin{minipage}[b]{0.24\linewidth}\centering\includegraphics[width=\linewidth]{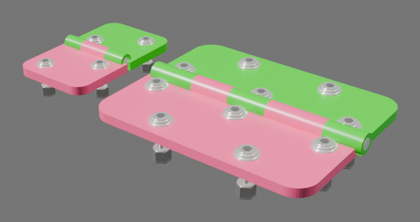}\\[1pt]{\scriptsize Round 10}\end{minipage}\hfill
    \begin{minipage}[b]{0.24\linewidth}\centering\includegraphics[width=\linewidth]{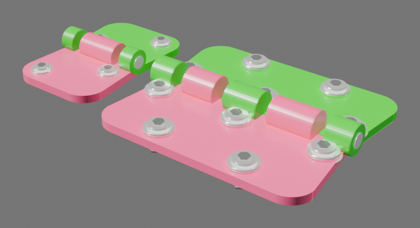}\\[1pt]{\scriptsize Round 22}\end{minipage}
    \caption{Reconstruction of the NopSCADlib flat hinge, shown from one oblique view. The evaluator accepted round 22. Its first review began: ``The main hinge leaf outline is substantially too regular and too large/deep compared with the source CSG leaf shape.''}
    \label{fig:recon-trajectory}
\end{figure}

\paragraph{OpenSCAD reconstruction.}
Converting OpenSCAD directly to Blender does not yield editable programs. Translating the constructive solid geometry keeps the shape but flattens modules and parameters into raw vertices, transformation matrices, and Boolean operations, so the program no longer exposes concepts such as a part count or a structural parameter. Prompting a frontier model once to rewrite the design keeps the structure but loses fidelity on complex objects. We therefore reconstruct each model with Codex CLI agents backed by GPT-5.5. In each round, a generator writes a Blender program, renders it, and compares the renders with those of the OpenSCAD source. An evaluator, which does not see the generator's reasoning, then returns concrete revision feedback. A coordinator agent alternates the two and stops when the evaluator reports that no meaningful geometric or structural defect remains, or when the round budget is exhausted. Reconstructions take 5.5 rounds on average. Prompts~\ref{prompt:recon-gen} and~\ref{prompt:recon-eval} abbreviate the generator and evaluator briefs, and Figure~\ref{fig:recon-trajectory} shows one of the longest trajectories.

\begin{promptbox}[label=prompt:recon-gen]{OpenSCAD reconstruction: generator brief (abbreviated)}
You are the generator for one \textasciigrave{}scad -{}\textgreater{} bpy\textasciigrave{} reconstruction job in the \textasciigrave{}scad\_\allowbreak{}bpy\_\allowbreak{}jobs\textasciigrave{} workspace.
\par
[...]
\par
This project reconstructs a single source OpenSCAD model as reusable,\allowbreak{} readable Blender Python. The goal is not only visual similarity. The goal is a maintainable BPY reconstruction with a clear split between low-{}level reusable geometry helpers and high-{}level assembly logic.
\par
The required code split is:
\par
-{} \textasciigrave{}primitive\_\allowbreak{}spec.py\textasciigrave{} for low-{}level primitives and reusable geometry helpers\newline
-{} \textasciigrave{}procedural\_\allowbreak{}construction.py\textasciigrave{} for readable composition of the final model\newline
-{} \textasciigrave{}render\_\allowbreak{}entry.py\textasciigrave{} as a minimal render entrypoint
\par
Focus primarily on structure and geometry. You may and should reconstruct visible source colors and basic material assignments when they are apparent,\allowbreak{} especially when colors distinguish model parts,\allowbreak{} but treat them as secondary to shape,\allowbreak{} proportions,\allowbreak{} placement,\allowbreak{} and reusable construction. Do not spend generation time on texture,\allowbreak{} lighting style,\allowbreak{} or photoreal material polish unless it clarifies geometry.
\par
\#\# Your Role
\par
You are the implementation agent for the current iteration. You are not the coordinator and you are not the reviewer.
\par
Your job is to produce the best possible implementation for the assigned current iteration,\allowbreak{} using all available project history as context.
\par
[...]
\par
You may run your own generation-{}refine loop inside the assigned iteration. For example:
\par
1. inspect source and prior history\newline
2. implement or revise BPY code\newline
3. render the current iteration\newline
4. compare against source renders\newline
5. refine again if needed
\par
Use good judgment,\allowbreak{} but keep the loop bounded and purposeful.
\par
[...]
\par
\#\# Output Expectations
\par
Keep \textasciigrave{}notes.md\textasciigrave{} current with concise implementation rationale,\allowbreak{} major geometric decisions,\allowbreak{} and open risks that the reviewer should pay attention to.
\par
When your implementation pass is done,\allowbreak{} leave the current iteration in a reviewable state:
\par
-{} BPY code exists and runs\newline
-{} canonical BPY renders exist for the iteration\newline
-{} \textasciigrave{}notes.md\textasciigrave{} explains the main approach
\par
Do not declare the job complete yourself. Completion is decided by the reviewer and coordinator.
\end{promptbox}

\begin{promptbox}[label=prompt:recon-eval]{OpenSCAD reconstruction: evaluator brief (abbreviated)}
You are the independent reviewer for one \textasciigrave{}scad -{}\textgreater{} bpy\textasciigrave{} reconstruction job in the \textasciigrave{}scad\_\allowbreak{}bpy\_\allowbreak{}jobs\textasciigrave{} workspace.
\par
[...]
\par
\#\# Your Role
\par
You are not the implementation agent. Your job is to inspect the current job state and write review findings to the output path given by the coordinator,\allowbreak{} usually the current iteration\textquotesingle{}s \textasciigrave{}feedback.md\textasciigrave{}.
\par
[...]
\par
\#\# Review Focus
\par
Prioritize:
\par
1. visible geometric mismatch\newline
2. missing parts,\allowbreak{} wrong proportions,\allowbreak{} wrong placement,\allowbreak{} wrong repetition\newline
3. poor reuse or a weak split between primitives and procedural construction\newline
4. the smallest set of high-{}value next edits
\par
Only evaluate structure and geometry. Ignore color,\allowbreak{} materials,\allowbreak{} textures,\allowbreak{} and lighting style unless they hide or clarify geometric evidence. Do not ask for another iteration for color-{}only or material-{}only differences.
\par
[...]
\par
If the current BPY result is already materially complete,\allowbreak{} say so plainly instead of inventing more critique.
\par
\#\# Output Rules
\par
Write the review to the requested \textasciigrave{}feedback.md\textasciigrave{} path instead of returning only chat output.
\par
Format the file like this:
\par
1. findings first,\allowbreak{} ordered by severity\newline
2. each finding cites the image names and code files used\newline
3. then a short next-{}edits list
\par
If the reconstruction is complete,\allowbreak{} write a completion verdict instead of normal findings. Use this exact sentence:
\par
\textasciigrave{}This reconstruction is complete. I recommend ending the job and marking this iteration as the final version.\textasciigrave{}
\par
Use that sentence only when there are no meaningful remaining defects.
\end{promptbox}

\begin{figure}[h]
    \centering
    
    \begin{minipage}[t]{0.49\linewidth}\centering\parbox[c][2.9cm][c]{\linewidth}{\centering\includegraphics[height=2.9cm,width=\linewidth,keepaspectratio]{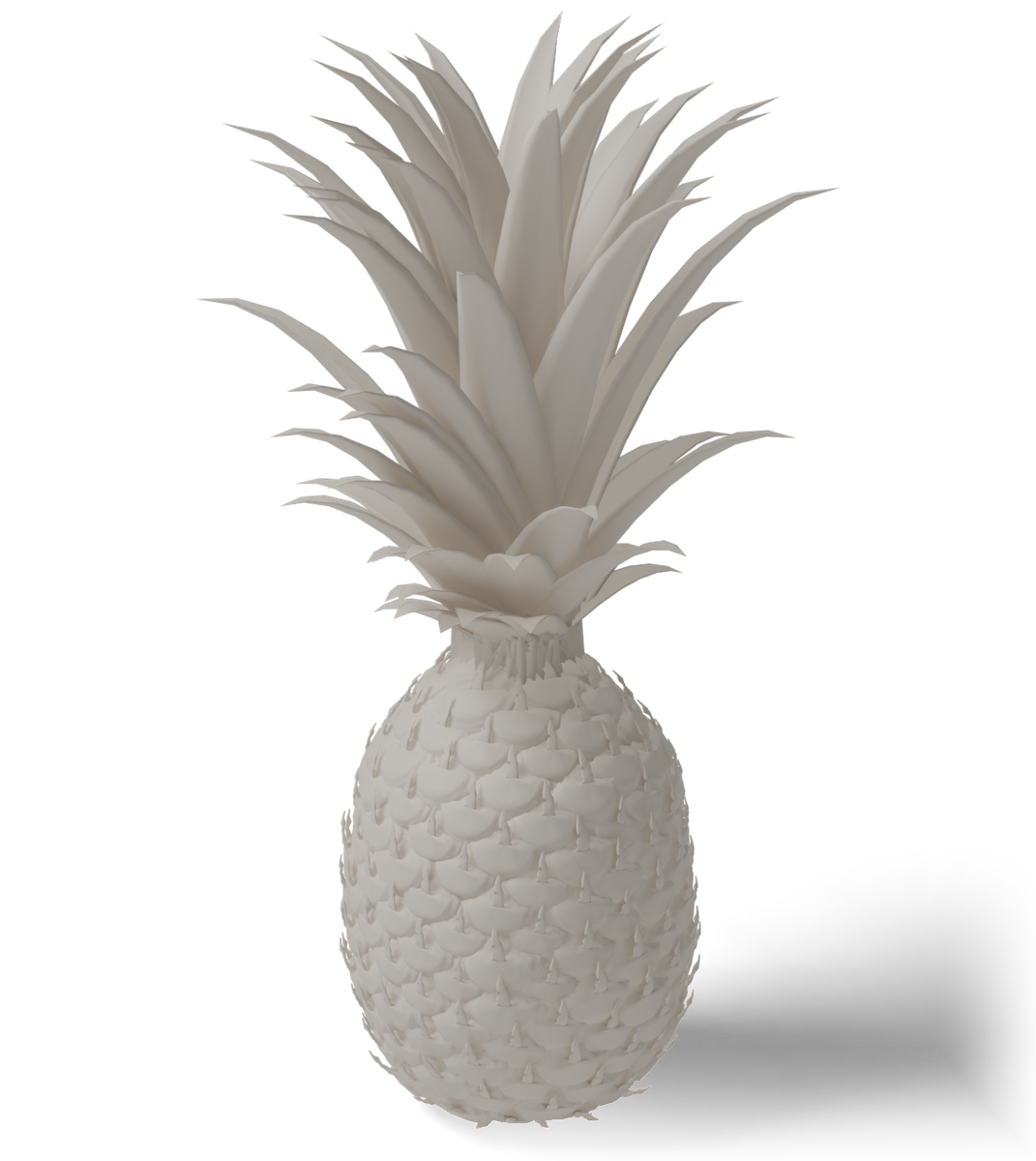}}\\[1pt]{\small 3DCodeBench}\end{minipage}\hfill
    \begin{minipage}[t]{0.49\linewidth}\centering\parbox[c][2.9cm][c]{\linewidth}{\centering\includegraphics[height=2.5cm,width=\linewidth,keepaspectratio]{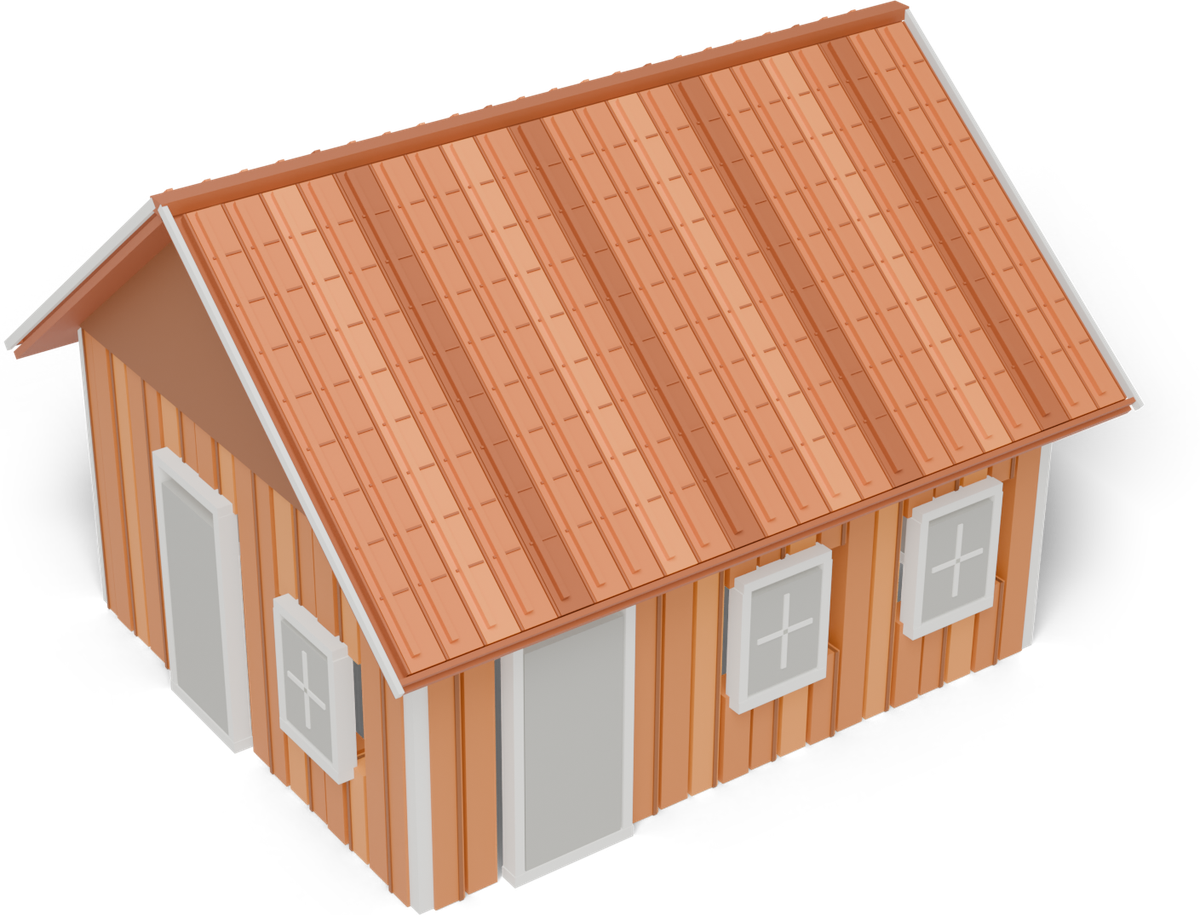}}\\[1pt]{\small Architecture}\end{minipage}\\[6pt]
    \begin{minipage}[t]{0.49\linewidth}\centering\parbox[c][2.9cm][c]{\linewidth}{\centering\includegraphics[height=2.5cm,width=\linewidth,keepaspectratio]{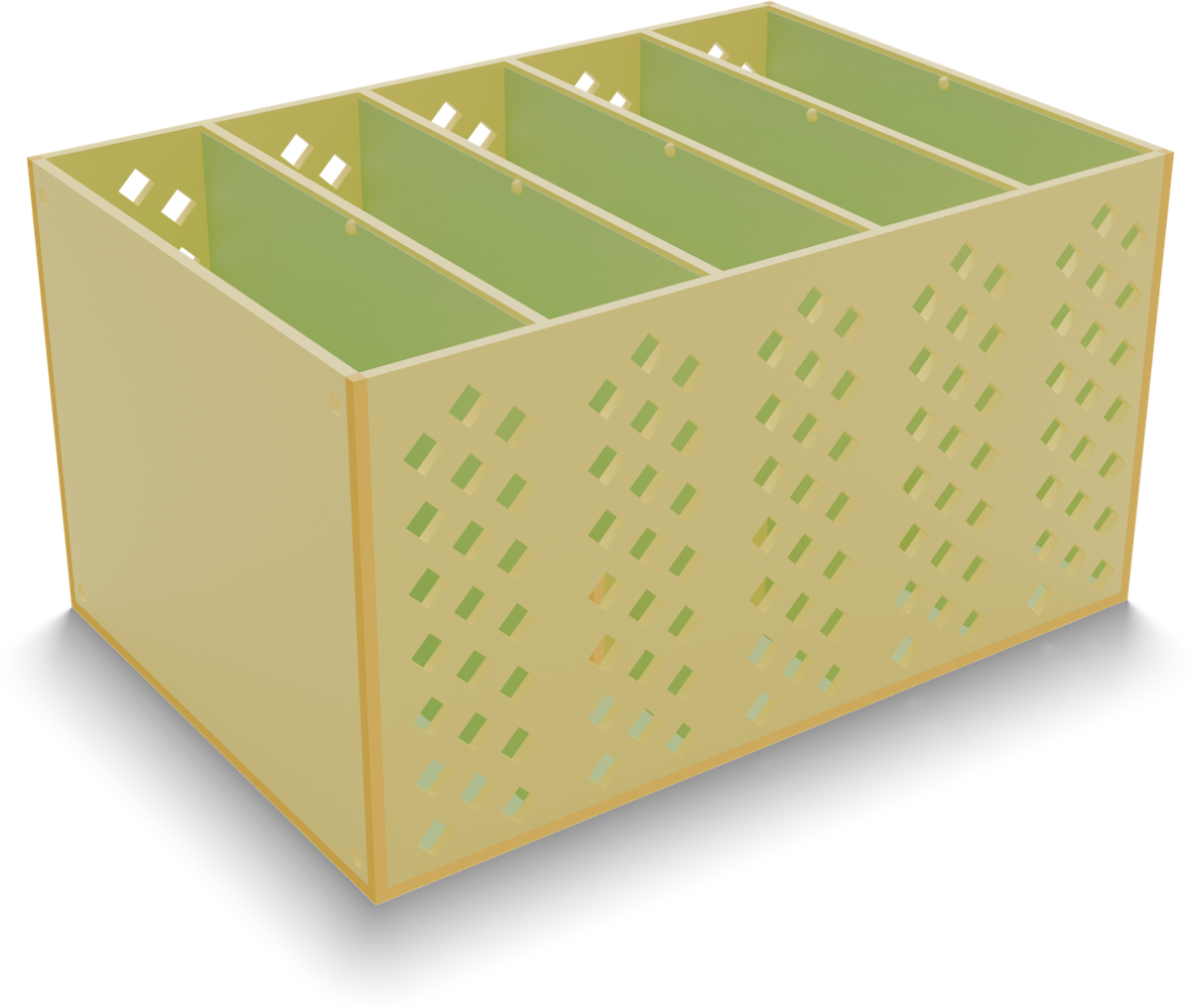}}\\[1pt]{\small Furniture}\end{minipage}\hfill
    \begin{minipage}[t]{0.49\linewidth}\centering\parbox[c][2.9cm][c]{\linewidth}{\centering\includegraphics[height=2.3cm,width=\linewidth,keepaspectratio]{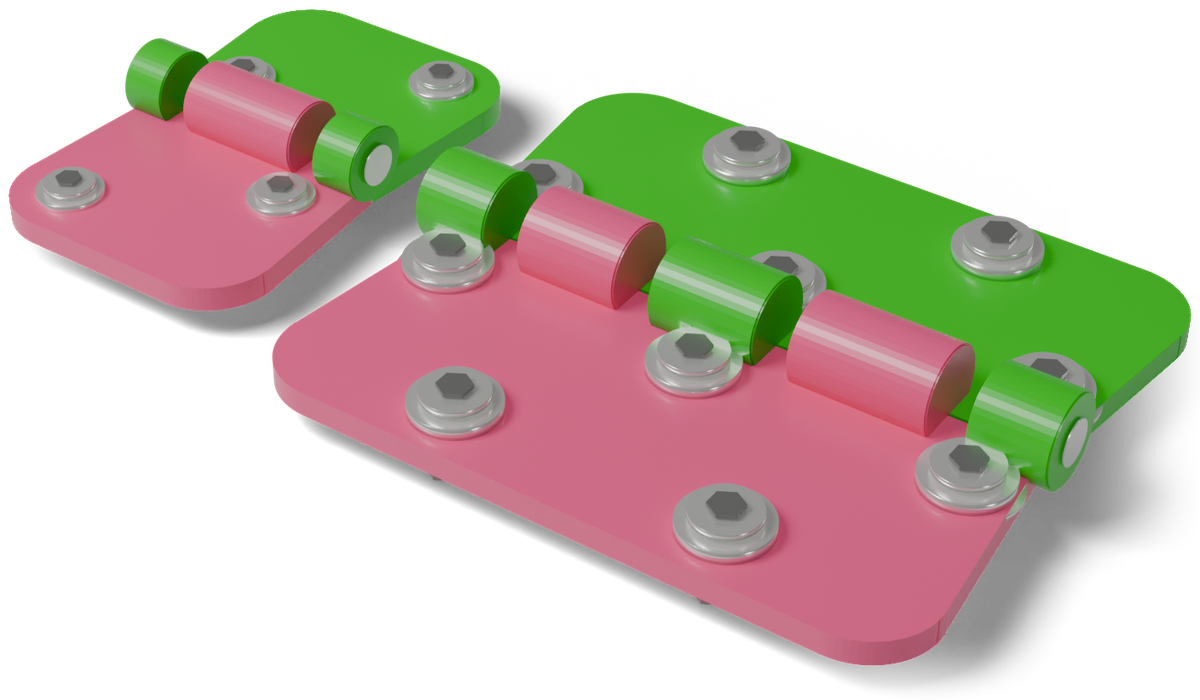}}\\[1pt]{\small Manufacturing}\end{minipage}
    \caption{\looseness=-1
    Example source objects from the four source groups of \method{}: a pineapple (546 lines of code) from 3DCodeBench, and a board-and-batten cottage (677), a small-parts cabinet (717), and a flat hinge (760) reconstructed from OpenSCAD, rendered from the source programs.}
    \label{fig:source-examples}
\end{figure}

\paragraph{Curation.}
From the 127 reconstructions, we remove relatively simple objects and objects whose type duplicates another reconstruction, and we exclude every model whose upstream license does not clearly permit redistributing a derived program. The 53 retained programs comprise 16 architecture, 19 furniture, and 18 manufacturing objects. Figure~\ref{fig:source-examples} shows one source object per category.

\paragraph{Licenses and intended use.}
The 55 3DCodeBench programs are distributed under BSD-3-Clause and MIT, as derived from Infinigen. The 53 reconstructed programs derive from OpenSCAD models released under CC BY 4.0 (16), CC BY-SA 4.0 (13), CC BY-NC 4.0 (7), GPL-3.0 (4), CC BY-NC-SA 4.0 (3), CC0 1.0 (3), MIT (3), and other open licenses (4). We release each program under its upstream license with attribution to its creator, our prompts and checklists under CC BY 4.0, and our code under the MIT license. \method{} is intended for research on evaluating code editing, and programs under non-commercial licenses may be used only for non-commercial purposes.

\begin{figure*}[t]
    \centering
    \includegraphics[width=\textwidth]{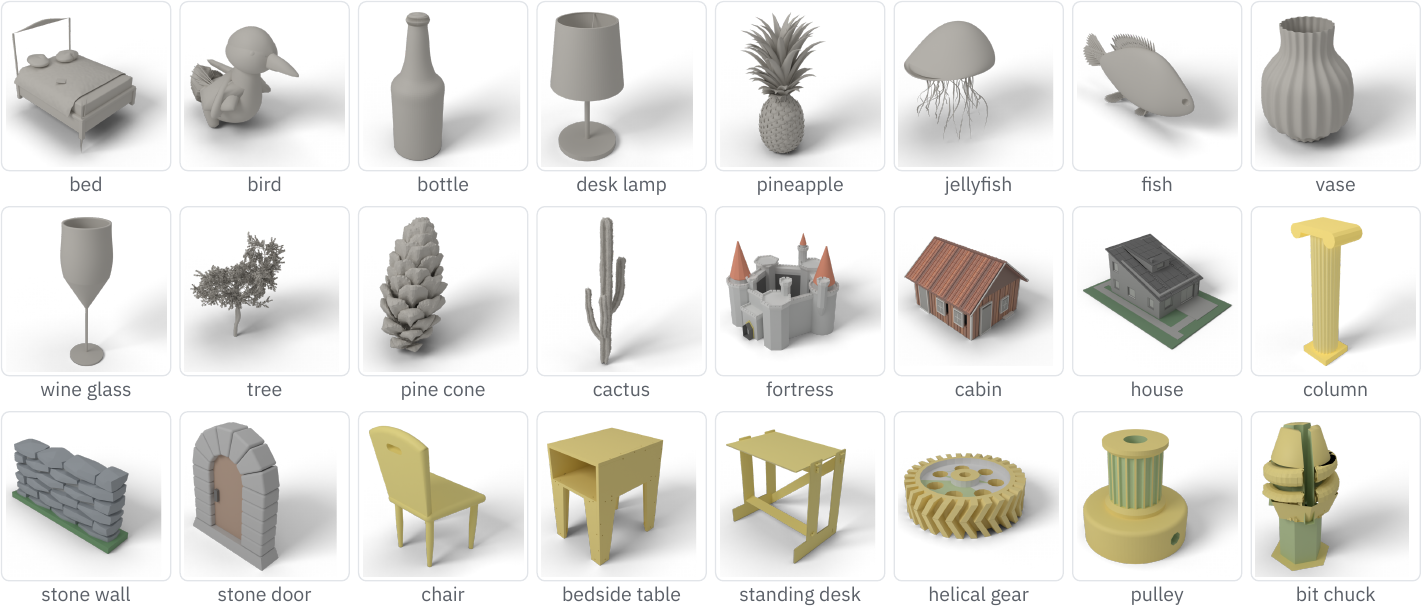}
    \caption{\looseness=-1
    Twenty-four of the 108 source programs: 3DCodeBench objects (grey) and reconstructed designs (coloured).}
    \label{fig:source-gallery}
\end{figure*}

\subsection{Prompt Generation}
\label{app:instructions}
\looseness=-1
\paragraph{Generation.}
For each program and edit type, GPT-5.6-sol receives the object's name and one-line description, the program, and four renders from the corners of the object ($512\times512$), together with the definition of the edit type and one worked example. It returns eight candidates, each an edit-only prompt with one checklist item per instruction. Prompt~\ref{prompt:gen} shows the system prompt, the user template, and the definition of G4. The prompts number the edit types as Levels 1--6 (G1--G4, S, C). All edit prompts and checklists are in English.

\begin{promptbox}[label=prompt:gen]{Prompt generation (GPT-5.6-sol)}
\pnote{System prompt}
\par
You design visually verifiable 3D editing benchmark cases. Given a Blender Python source,\allowbreak{} its description,\allowbreak{} and four labeled corner renders,\allowbreak{} generate modification queries and aligned evaluation checklists at the requested difficulty level.
\par
Inspect the source and all four labeled corner renders before proposing an edit. Each generated query clause must describe an edit result that is directly visible,\allowbreak{} clearly false in the original,\allowbreak{} and recognizable at the supplied image scale. Generate only the level-{}specific edit clauses and checklist items. The framework prepends one standardized source-{}preservation clause and gate after generation for every level.
\par
Design every result for native 512 x 512 full-{}object renders. Its defining silhouette,\allowbreak{} extent,\allowbreak{} spacing,\allowbreak{} or orientation must occupy a substantial visible span and remain unmistakable without cropping or enlargement. Localized edits need clearly separated boundaries at that scale. Use approximate angles only when the coarse directional change is visually obvious rather than dependent on a few pixels of contour.
\par
An eligible target is a visually identifiable object,\allowbreak{} exterior region,\allowbreak{} or coherent part family. Use source code to establish the real baseline and a plausible edit,\allowbreak{} while writing the query and checklist entirely in visible object language. Use only object-{}relative roles and relationships that remain stable across camera directions. Do not define a target,\allowbreak{} direction,\allowbreak{} orientation,\allowbreak{} or outcome by image or screen position,\allowbreak{} a named view,\allowbreak{} or its relation to the camera.
\par
Checklist evidence is limited to rendered pixels. Exclude names,\allowbreak{} metadata,\allowbreak{} implementation methods,\allowbreak{} topology or connectivity state,\allowbreak{} hidden or internal geometry,\allowbreak{} underside-{}only features,\allowbreak{} and details too small to judge reliably. Material or lighting cues are eligible only when a level explicitly allows a broad,\allowbreak{} high-{}contrast visible result. Exact counts are eligible when the required members can be distinguished reliably across the supplied views.
\par
Write one semicolon-{}delimited query clause for each checklist item,\allowbreak{} in the same order. Each clause/\allowbreak{}item pair must state one complete visible outcome and preserve every count,\allowbreak{} shape,\allowbreak{} size,\allowbreak{} placement,\allowbreak{} identity,\allowbreak{} scope,\allowbreak{} and relationship qualifier. Checklist items are concise yes/\allowbreak{}no questions and contain no camera,\allowbreak{} image-{}edge,\allowbreak{} screen-{}position,\allowbreak{} or view-{}selection language. All requested outcomes must coexist and remain independently judgeable.
\par
\pnote{User message (followed by the four corner renders)}
\par
\#\# Object\newline
-{} Name: \{name\}\newline
-{} Description: \{instruction\}
\par
\#\# Blender Script\newline
\mbox{}\textasciigrave{}\textasciigrave{}\textasciigrave{}python\newline
\mbox{}\{script\}\newline
\mbox{}\textasciigrave{}\textasciigrave{}\textasciigrave{}
\par
\#\# Attached Evidence\newline
Four labeled original renders are attached: front\_\allowbreak{}left,\allowbreak{} front\_\allowbreak{}right,\allowbreak{} back\_\allowbreak{}left,\allowbreak{} and back\_\allowbreak{}right.
\par
\#\# Task\newline
Generate exactly \{n\} modification queries at Level \{level\}.
\par
\{level\_\allowbreak{}description\}
\par
Return a JSON array with exactly \{n\} elements:\newline
\mbox{}\textasciigrave{}\textasciigrave{}\textasciigrave{}json\newline
\mbox{}[\newline
\mbox{}~~\{\newline
\mbox{}~~~~"query": "semicolon-{}delimited modification clauses",\allowbreak{}\newline
\mbox{}~~~~"checklist": ["aligned yes/\allowbreak{}no checklist item",\allowbreak{} "..."]\newline
\mbox{}~~\}\newline
\mbox{}]\newline
\mbox{}\textasciigrave{}\textasciigrave{}\textasciigrave{}\newline
Return only the JSON array.
\par
\pnote{\{level\_description\} for Level 4 (G4, Structural). Levels 1--3, 5, and 6 (G1--G3, S, C) use the same format; Table~\ref{tab:taxonomy} summarizes them.}
\par
**Level 4 --- Coordinated visible structural reconstruction**\newline
Reorganize how at least three identifiable regions of the existing object meet,\allowbreak{} span,\allowbreak{} support,\allowbreak{} repeat,\allowbreak{} or transition into one another. Create one coordinated structural reconstruction rather than a collection of unrelated local edits.
\par
Request five independently judgeable outcomes. Together they cover a new large-{}scale layout or silhouette,\allowbreak{} at least two changed region-{}to-{}region relationships,\allowbreak{} and additional broad form,\allowbreak{} opening,\allowbreak{} spacing,\allowbreak{} or proportion results. State visible spans,\allowbreak{} gaps,\allowbreak{} alignments,\allowbreak{} openings,\allowbreak{} transitions,\allowbreak{} and silhouettes rather than implementation or physical-{}connectivity claims.
\par
Example:\newline
-{} Query: "Rebuild the bracket as two outward-{}tapering arms joined by a broad triangular corner web; cut a large rounded opening through the web; form two thick ribs spanning from the web onto the arms; splay the arm ends farther apart; make the reconstructed corner region conspicuously taller."\newline
-{} Checklist: "Do the two arms taper outward from a broad triangular corner web?" \& "Is there a large rounded opening through the web?" \& "Are two thick ribs visibly spanning from the web onto the arms?" \& "Are the arm ends conspicuously farther apart?" \& "Is the reconstructed corner region conspicuously taller than in the original?"
\par
Write exactly five query clauses and five checklist items.
\end{promptbox}

\looseness=-1
\paragraph{Admission and human review.}
The admission model rejects a candidate whose evidence depends on names, metadata, implementation method, topology or connectivity, hidden or internal geometry, subtle surface appearance, or details below reliable image scale; whose baseline is unsupported by the source; whose target is ambiguous among repeated parts; whose outcomes are bundled or duplicated; whose checklist drifts from its instructions; that violates the edit-type definition; or that refers to image position, a named view, or the camera. Human annotators then review the admitted candidates and reject those that are ambiguous, not visually verifiable, or unreasonable. All human review and annotation in this paper, including this review and the labels used to validate the verifier (Appendix~\ref{app:verifier}) and the error analysis (Appendix~\ref{app:errors}), were done by the authors. Of 5,620 candidates for the 108 programs, 4,123 (73.4\%) pass both filters. Prompt~\ref{prompt:admit} gives the admission rubric. If fewer than four candidates of a program and edit type pass, more are generated.

\begin{promptbox}[label=prompt:admit]{Candidate admission (GPT-5.6-sol)}
\pnote{System prompt}
\par
You audit candidate queries for a visually grounded 3D editing benchmark. Use only the canonical source record and the four original corner renders. Decide whether each candidate satisfies the supplied level contract and can be scored reliably from rendered pixels. Return only the requested JSON.
\par
\pnote{User message; \{admission\_rubric\} is given below}
\par
Audit every candidate against this rubric:
\par
\{admission\_\allowbreak{}rubric\}
\par
\#\# Source metadata\newline
\{source\_\allowbreak{}metadata\}
\par
\#\# Canonical source\newline
\mbox{}\textasciigrave{}\textasciigrave{}\textasciigrave{}python\newline
\mbox{}\{script\}\newline
\mbox{}\textasciigrave{}\textasciigrave{}\textasciigrave{}
\par
\#\# Candidates\newline
\{candidates\_\allowbreak{}json\}
\par
The four labeled original renders are attached after this message.
\par
Return a JSON array with one object per candidate,\allowbreak{} in the supplied order:\newline
\mbox{}\textasciigrave{}\textasciigrave{}\textasciigrave{}json\newline
\mbox{}[\newline
\mbox{}~~\{\newline
\mbox{}~~~~"candidate\_\allowbreak{}id": "...",\allowbreak{}\newline
\mbox{}~~~~"admission": "ACCEPT or REJECT",\allowbreak{}\newline
\mbox{}~~~~"reason\_\allowbreak{}codes": ["..."],\allowbreak{}\newline
\mbox{}~~~~"note": "brief source-{}grounded explanation"\newline
\mbox{}~~\}\newline
\mbox{}]\newline
\mbox{}\textasciigrave{}\textasciigrave{}\textasciigrave{}
\par
\pnote{\{admission\_rubric\}}
\par
Admit a candidate when all of the following hold:\newline
1. Every generated edit clause states a visible result that is clearly false in the original. The candidate shown for audit already includes the binding framework-{}owned source-{}preservation pair.\newline
2. Every target and relationship is unambiguous from its object-{}relative role,\allowbreak{} region,\allowbreak{} landmarks,\allowbreak{} or coherent-{}family scope.\newline
3. Every result is large,\allowbreak{} exposed,\allowbreak{} and directly judgeable across the four native 512 x 512 full-{}object renders without cropping or enlargement. Its defining edges,\allowbreak{} silhouette,\allowbreak{} extent,\allowbreak{} spacing,\allowbreak{} or coarse direction occupy enough pixels for a stable yes/\allowbreak{}no comparison.\newline
4. Query clauses and checklist items align one-{}to-{}one and preserve every explicit qualifier.\newline
5. A successful edit would remain a recognizable continuation of the original object. Apply this strictly for L1-{}L4,\allowbreak{} broadly at the underlying-{}category level for L5,\allowbreak{} and as visible original-{}object presence within the scene for L6.
\par
Reject a candidate whose evidence depends on names,\allowbreak{} metadata,\allowbreak{} implementation method,\allowbreak{} topology or connectivity,\allowbreak{} hidden/\allowbreak{}internal/\allowbreak{}underside geometry,\allowbreak{} subtle surface appearance,\allowbreak{} or details below reliable image scale. Also reject an unsupported baseline,\allowbreak{} ambiguous repeated target,\allowbreak{} bundled or duplicated outcome,\allowbreak{} clause/\allowbreak{}checklist drift,\allowbreak{} or violation of the level contract. Reject any candidate that defines a target,\allowbreak{} direction,\allowbreak{} orientation,\allowbreak{} or outcome by image or screen position,\allowbreak{} a named view,\allowbreak{} or its relation to the camera instead of stable object-{}relative landmarks. Treat a shape class,\allowbreak{} ratio,\allowbreak{} or directional claim that depends on a few pixels of contour as BELOW\_\allowbreak{}SCALE or OUTCOME\_\allowbreak{}AMBIGUOUS rather than assuming that a higher-{}resolution render would resolve it.
\par
Level contracts:
\par
[...]
\end{promptbox}

\looseness=-1
\paragraph{Selection.}
GPT-5.6-sol then keeps four candidates per program and edit type (Prompt~\ref{prompt:select}). It ranks them by how clearly the results would be visible, how natural and well aligned the requests are, and how diverse their targets are, without seeing any editor output.

\begin{promptbox}[label=prompt:select]{Candidate selection (GPT-5.6-sol, abbreviated)}
\pnote{System prompt}
\par
You freeze a target-{}blind query bank for a 3D editing benchmark. Select only from source-{}admitted candidates using the canonical source and four original corner renders. Rank source-{}side visual quality and diversity without using target-{}model outputs or expected target performance. Return only the requested JSON.
\par
\pnote{User message}
\par
Select \{selection\_\allowbreak{}count\} candidates for each requested level.
\par
Prioritize,\allowbreak{} in order:\newline
1. Direct,\allowbreak{} unmistakable visibility of every target and claimed result across the four native 512 x 512 full-{}object renders,\allowbreak{} with a robust visible span and boundary at that scale.\newline
2. Natural,\allowbreak{} coherent requests with clean one-{}to-{}one clause/\allowbreak{}checklist alignment.\newline
3. Diverse targets and edit types within each level.\newline
4. Faithfulness to these level contracts:\newline
\{level\_\allowbreak{}contracts\}
\par
The supplied candidates contain edit outcomes only. Account for the framework-{}owned source-{}preservation gate when assessing whether the completed result would retain the appropriate identity: strict same-{}object recognition for L1-{}L4,\allowbreak{} broad underlying-{}category recognition for L5,\allowbreak{} and visible original-{}object presence for L6.
\par
[...]
\par
The four labeled original renders are attached after this message.
\par
Return a JSON array ordered by level and selection rank:\newline
\mbox{}\textasciigrave{}\textasciigrave{}\textasciigrave{}json\newline
\mbox{}[\newline
\mbox{}~~\{\newline
\mbox{}~~~~"candidate\_\allowbreak{}id": "...",\allowbreak{}\newline
\mbox{}~~~~"level": 1,\allowbreak{}\newline
\mbox{}~~~~"selection\_\allowbreak{}rank": 1,\allowbreak{}\newline
\mbox{}~~~~"rationale": "brief concrete reason"\newline
\mbox{}~~\}\newline
\mbox{}]\newline
\mbox{}\textasciigrave{}\textasciigrave{}\textasciigrave{}
\end{promptbox}

\subsection{Ambiguity Screening}
\label{app:ambiguity}
\looseness=-1
We screen each checklist item with Claude Opus 5.5 in two separate calls, one given the prompt and checklist only and one also given the source program, and flag the item if either call does. For each flagged task, a resolver call decides for each flagged item whether to keep it, revise it, or drop it. A rewriter then revises only the flagged instructions and their partners, keeping the requested difficulty, and the revised task is screened again, with up to two rewriting rounds. All rewrites are made under human supervision. The process revises 787 of the 2,592 tasks, and the 51 items (of 12,528) that remain ambiguous are excluded from scoring. Prompt~\ref{prompt:screen} is the screening prompt, and Prompts~\ref{prompt:resolve} and~\ref{prompt:rewrite} abbreviate the resolver and the rewriter. Table~\ref{tab:ambiguity-examples} shows two rewritten instructions: ``about 50\% larger'' can refer to the diameter or the area of each opening, and a pedestal-mounted arm conflicts with a gantry that carries it.

\begin{table}[t]
\centering
\footnotesize
\begin{tabular}{@{}p{0.97\columnwidth}@{}}
\toprule
\textbf{Helical gear, G3, term}\\
\emph{Before:} enlarge each of the seven spiral-arranged openings by about 50\%\\
\emph{After:} enlarge each of the seven spiral-arranged openings to about 1.5 times its original diameter\\
\midrule
\textbf{Pulley scene, C, conflict between instructions}\\
\emph{Before:} add a pedestal-mounted two-link robotic arm with a large two-finger gripper reaching toward the pulley\\
\emph{After:} add a two-link robotic arm rising from a pedestal block fixed to the gantry, with a large two-finger gripper reaching toward the pulley\\
\emph{Conflicting instruction:} build a large open gantry spanning the workbench, carrying the robotic arm (after: the robotic arm's pedestal block) above the turntable [...]\\
\bottomrule
\end{tabular}
\caption{\looseness=-1
Two instructions before and after ambiguity rewriting. Their checklist items were rewritten in the same way.}
\label{tab:ambiguity-examples}
\end{table}

\begin{promptbox}[label=prompt:screen]{Ambiguity screening (Claude Opus 5.5, abbreviated)}
\pnote{System prompt}
\par
You audit checklist items from a benchmark of 3D-{}model code edits. A model edits a Blender Python script so the rendered object satisfies an edit instruction. A verifier then checks each checklist item against six rendered views (front,\allowbreak{} back,\allowbreak{} left,\allowbreak{} right,\allowbreak{} top,\allowbreak{} bottom) of the ORIGINAL and the MODIFIED object.
\par
Task: decide whether the TARGET checklist item is AMBIGUOUS.
\par
An item is ambiguous if a careful,\allowbreak{} competent judge could defensibly read it in two different ways,\allowbreak{} such that a plausible edited object would PASS under one reading and FAIL under the other. Types:\newline
-{} frame: a direction or dimension word ("side to side",\allowbreak{} "front",\allowbreak{} "rear",\allowbreak{} "up",\allowbreak{} "height",\allowbreak{} "width",\allowbreak{} "tallest",\allowbreak{} "depth") can refer to different axes. Examples: the object\textquotesingle{}s own frame vs the rendered or viewer frame,\allowbreak{} or a part\textquotesingle{}s local frame vs the whole object. The two readings point to different geometry.\newline
-{} reference: it is unclear which part,\allowbreak{} region or feature is meant (e.g. "the tail base",\allowbreak{} "the outer sector",\allowbreak{} "the original cushion"),\allowbreak{} and the candidates differ.\newline
-{} term: an interpretive shape or style word ("spear-{}like",\allowbreak{} "three-{}lobed",\allowbreak{} "gnarled",\allowbreak{} "dragon-{}like") has materially different reasonable senses.\newline
-{} conflict: the instruction\textquotesingle{}s clauses contradict each other,\allowbreak{} or the edit contradicts the preservation item,\allowbreak{} so no edit can satisfy every reading.\newline
-{} other: a different genuine ambiguity (explain).
\par
These do NOT make an item ambiguous:\newline
-{} ordinary vague magnitude words ("conspicuously",\allowbreak{} "approximately twice",\allowbreak{} "broad"),\allowbreak{} as long as the intended direction and target are clear;\newline
-{} an item that is hard to satisfy or hard to verify but has one natural reading.
\par
Decide from the information you are given. Say what you relied on.
\par
Return only one JSON object:\newline
\{"ambiguous": true|false,\allowbreak{} "type": "frame|reference|term|conflict|other|none",\allowbreak{} "reading\_\allowbreak{}a": "one reading (or empty)",\allowbreak{} "reading\_\allowbreak{}b": "a competing reading (or empty)",\allowbreak{} "needs": "text|original\_\allowbreak{}render|edited\_\allowbreak{}render|code",\allowbreak{} "confidence": 0.0-{}1.0,\allowbreak{} "rationale": "one or two sentences"\}
\par
[...]
\par
\pnote{User message; the second call appends the program}
\par
OBJECT SOURCE: \{entry\_\allowbreak{}id\} (level L\{level\})\newline
EDIT INSTRUCTION:\newline
\{instruction\}
\par
FULL CHECKLIST:\newline
\{checklist\}
\par
TARGET ITEM: [\{index\}] \{item\}
\par
ORIGINAL PROGRAM (Blender Python):\newline
\mbox{}\textasciigrave{}\textasciigrave{}\textasciigrave{}python\newline
\mbox{}\{program\}\newline
\mbox{}\textasciigrave{}\textasciigrave{}\textasciigrave{}
\end{promptbox}

\begin{promptbox}[label=prompt:resolve]{Flag resolution (Claude Opus 5.5, abbreviated)}
You review flags raised by an automatic screener on the checklist of ONE task from a benchmark of 3D-{}model code edits.
\par
[...]
\par
The screener examined each item separately and flagged some as ambiguous,\allowbreak{} giving two competing readings. Some flags are real and some are false alarms. For every FLAGGED item,\allowbreak{} decide whether to KEEP it,\allowbreak{} REVISE it,\allowbreak{} or DROP it. Unflagged items are not your concern,\allowbreak{} except as context.
\par
[...]
\par
\#\# Decision rules\newline
1. KEEP (a false alarm) when the screener\textquotesingle{}s competing reading is strained,\allowbreak{} when both readings give the same verdict for plausible edits,\allowbreak{} when the only issue is an ordinary vague magnitude word ("conspicuously",\allowbreak{} "approximately twice") with a clear direction and target,\allowbreak{} or when an apparent conflict disappears under a natural reading that satisfies both items at once.
\par
[...]
\par
4. REVISE when a short rewording removes the ambiguity without changing what the instruction asks for. Give the rewritten item in \textasciigrave{}rewrite\textasciigrave{}; it must be decidable from the six views,\allowbreak{} must not add requirements,\allowbreak{} and should follow the instruction\textquotesingle{}s wording where possible.\newline
5. DROP when the intended meaning cannot be recovered from the instruction (a rewrite would have to guess).
\par
[...]
\par
\#\# Output\newline
Return only one JSON object,\allowbreak{} with one entry per FLAGGED item,\allowbreak{} in the order given:\newline
\{"items": [\{"index": 3,\allowbreak{} "decision": "keep|revise|drop",\allowbreak{} "issue": "none|frame|reference|term|conflict|other",\allowbreak{} "conflicts\_\allowbreak{}with": [],\allowbreak{} "reading\_\allowbreak{}a": "",\allowbreak{} "reading\_\allowbreak{}b": "",\allowbreak{} "rewrite": "",\allowbreak{} "rationale": ""\}]\}
\par
[...]
\end{promptbox}

\begin{promptbox}[label=prompt:rewrite]{Rewriting flagged instructions (Claude Opus 5.5, abbreviated)}
You revise ONE task of a benchmark of 3D-{}model code edits so that it is unambiguous while keeping its difficulty.
\par
[...]
\par
\#\# Your job\newline
Rewrite the flagged clauses (and,\allowbreak{} where a flagged item conflicts with another clause,\allowbreak{} that partner clause) and their checklist items so that every requested outcome has exactly one natural reading.
\par
\#\# Rules\newline
1. Scope: change ONLY the flagged clauses/\allowbreak{}items and the clauses/\allowbreak{}items named as their conflict partners. Copy every other clause and item verbatim. Never change clause 0 or item 0. Keep the number and order of clauses and items,\allowbreak{} one item per clause.\newline
2. Never introduce colour,\allowbreak{} material or lighting requirements,\allowbreak{} and never change the direction of a comparison or a target relation (e.g. "nearly matches the larger claw" must not become "longer than the other claw"; "clamped to the workbench" must stay on the workbench).\newline
3. Keep the difficulty. Each rewritten clause must request the same outcome at the same scale: same targets,\allowbreak{} same count of outcomes,\allowbreak{} same magnitude words ("conspicuously",\allowbreak{} "approximately twice"),\allowbreak{} no requirement removed,\allowbreak{} weakened,\allowbreak{} merged or split. Add only the words needed to disambiguate; do not add new requirements.
\par
[...]
\par
6. Be concise. Add the fewest words that remove the ambiguity: one short anchoring phrase is usually enough. A rewritten clause should normally be at most about 1.5 times the length of the original. Do not describe the original object\textquotesingle{}s current state,\allowbreak{} do not add parenthetical explanations,\allowbreak{} and do not hint at how to implement the edit. Each checklist item is ONE yes/\allowbreak{}no question sentence that mirrors its clause.
\par
[...]
\par
\pnote{The system prompt ends with the shared grounding rules of Prompt~\ref{prompt:gen}. The user message gives the program, the level definition, the task, and the flags with their competing readings.}
\end{promptbox}

\section{Program Length of Blender Code Benchmarks}
\label{app:code-length}

\looseness=-1
Table~\ref{tab:code-length} reports the program length behind the ``Complex code'' column of Table~\ref{tab:benchmark-comparison}; a benchmark is marked complex when its programs average more than 200 lines. Length is the number of physical lines in each Python file, including comments and blank lines. For editing benchmarks we measure the program given to the model; for generation benchmarks we measure the reference program when one exists.

\begin{table}[h]
\centering
\small
\setlength{\tabcolsep}{3pt}
\begin{tabular}{llrrr}
\toprule
Benchmark & Programs & $N$ & Mean & Med. \\
\midrule
CADBench$^a$ & BlendNet & 12{,}059 & 38 & 36 \\
3DCodeBench & Reference & 212 & 476 & 363 \\
3DHarnessBench & Reference & 100 & 876 & 784 \\
Nova3D-Bench$^b$ & Generated & 54 & 489 & 402 \\
BlenderGym$^c$ & Start & 280 & 51 & 40 \\
BlenderBench & Start & 27 & 25 & 24 \\
\midrule
\method{} & Source & 108 & 613 & 440 \\
\quad 3DCodeBench & & 55 & 527 & 365 \\
\quad OpenSCAD recon. & & 53 & 703 & 538 \\
\bottomrule
\end{tabular}

\vspace{2pt}
\parbox{\linewidth}{\looseness=-1\footnotesize
$^a$CADBench releases no programs; we use BlendNet, the training set released with it \citep{du2024blenderllm}.
$^b$Nova3D-Bench releases no reference or start programs; we use the programs generated by Nova3D on its 54 items, as released by the authors.
$^c$Full public release (280 instances); the paper reports 245.}
\caption{Program length (lines) of Blender code benchmarks. Med.: median.}
\label{tab:code-length}
\end{table}

\section{Evaluation Protocol Details}
\label{app:protocol}
\looseness=-1
Each editor receives the task once, with reasoning effort medium (or an equivalent thinking budget for models without effort levels) and at most 64k output tokens, and a truncated or refused reply counts as a failed edit. We extract the edited program from each reply as the longest complete fenced code block that compiles, or otherwise as the reply with leading and trailing prose removed. Edited programs run in Blender~\citep{blender} 5.1.0 with a 480-second limit for execution and rendering. Each artifact is rendered from six canonical views at $512\times512$ with Cycles, and the original and edited artifacts share one camera framing, so that changes in size and position remain visible. A program fails if it raises an error, produces an empty scene, or exceeds the time limit. The verifier receives the six original and six edited views, the full prompt, and one checklist item per call, and answers yes or no; it does not see either program. Prompt~\ref{prompt:editor} is the single-shot editor prompt, and Prompt~\ref{prompt:verifier} is the verifier prompt. Gemini 3.8 Flash is called once per item with thinking level high.

\paragraph{Models.}
The main evaluation covers Claude Opus 5.5~\citep{anthropic2026claudeopus55}, Claude Sonnet 5.5~\citep{anthropic2026claudesonnet55}, Claude Haiku 5.5~\citep{anthropic2026claudehaiku55}, Claude Haiku 4.5~\citep{anthropic2025claudehaiku45}, GPT-6.1-sol~\citep{openai2026gpt61sol}, GPT-6-luna~\citep{openai2026gpt6astra}, Gemini 4 Argon~\citep{google2026gemini4argon}, Gemini 3.8 Flash~\citep{google2026gemini38flash}, Gemini 3.1 Pro~\citep{google2026gemini31pro}, Kimi K3~\citep{moonshot2026kimik3}, gpt-oss-120b~\citep{openai2025gptoss}, Gemma 4 31B~\citep{gemmateam2026gemma4}, Qwen3.8-27B, Qwen3.6-27B, Qwen3.6-35B-A3B, the Qwen3.5 models, Qwen3-VL-235B-A22B~\citep{qwen2025qwen3vl}, and Qwen3-Coder-Next~\citep{qwen2026qwen3codernext}. The timeline (Table~\ref{tab:timeline-models}) adds earlier Claude models, namely Sonnet 4~\citep{anthropic2025claude4}, Sonnet 4.5~\citep{anthropic2025claudesonnet45}, Opus 4.5~\citep{anthropic2025claudeopus45}, Opus 4.6~\citep{anthropic2026claudeopus46}, Sonnet 4.6~\citep{anthropic2026claudesonnet46}, Opus 4.8~\citep{anthropic2026claudeopus48}, Sonnet 5~\citep{anthropic2026claudesonnet5}, Opus 5~\citep{anthropic2026claudeopus5}, and Fable 5.1~\citep{anthropic2026claudefable51}. It also adds GPT-5.4~\citep{openai2026gpt54}, GPT-5.5~\citep{openai2026gpt55}, GPT-5.6-sol and GPT-5.6-luna~\citep{openai2026gpt56}, and GPT-6-astra~\citep{openai2026gpt6astra}, as well as Gemini 2.5 Flash~\citep{comanici2025gemini25}, Gemini 3 Flash~\citep{google2025gemini3flash}, Gemini 3.5 Flash~\citep{google2026gemini35flash}, Gemini 3.6 Flash~\citep{google2026gemini36flash}, Gemini 3.7 Flash~\citep{google2026gemini37flash}, Qwen2.5-Coder-32B~\citep{hui2024qwen25coder}, QwQ-32B~\citep{qwen2025qwq32b}, and Qwen3-32B~\citep{qwen2025qwen3}. The verifier comparison (Table~\ref{tab:verifier-accuracy}) further includes GPT-5.6-terra~\citep{openai2026gpt56}, Gemma 4 26B-A4B~\citep{gemmateam2026gemma4}, Grok 4.6~\citep{xai2026grok46}, and Kimi K2.5~\citep{kimiteam2026kimik25}. GPT-5.6-sol writes the candidate prompts, Claude Opus 5.5 screens them for ambiguity, and Gemini 3.8 Flash is the verifier. Open-weight models are served locally with vLLM~\citep{kwon2023vllm}.

Example~\ref{ex:armchair} and Figure~\ref{fig:armchair-example} show the G4 task of Figure~\ref{fig:teaser}c with the verdicts for two editors. GPT-6.1-sol passes every item. Gemma 4 31B raises the seat on four posts but builds straight runners, a flat back panel, and no arm rails, so items 3--5 fail. For item 4, the verifier cites the left view: ``The modified left view shows a flat, rectangular back panel rather than an arched one.''

\begin{promptbox}[label=prompt:editor]{Single-shot editor}
\pnote{System prompt}
\par
You are an expert Blender Python programmer. Given a base script and a modification instruction,\allowbreak{} return the complete modified script implementing every requested clause. Preserve unrelated behavior,\allowbreak{} definitions,\allowbreak{} imports,\allowbreak{} helpers,\allowbreak{} and geometry. Make no additional visible edits beyond those required by the instruction.
\par
Return only Python code.
\par
\pnote{User message}
\par
\#\# Base Script\newline
\mbox{}\textasciigrave{}\textasciigrave{}\textasciigrave{}python\newline
\mbox{}\{script\}\newline
\mbox{}\textasciigrave{}\textasciigrave{}\textasciigrave{}
\par
\#\# Modification Instruction\newline
\{query\}
\par
Return the complete modified Blender Python script.
\end{promptbox}

\begin{promptbox}[label=prompt:verifier]{Verifier (Gemini 3.8 Flash), one call per item}
\pnote{System prompt}
\par
You are a visual evaluation judge for 3D rendered objects. You receive six matched ORIGINAL/\allowbreak{}MODIFIED view pairs,\allowbreak{} the full modification query,\allowbreak{} one aligned query clause,\allowbreak{} and one checklist item. Inspect all six pairs and judge only the current checklist item from directly visible evidence.
\par
Use the full query only to understand intended edits,\allowbreak{} target roles,\allowbreak{} and which parts are targeted versus non-{}targeted. The query is not evidence that an edit succeeded. Do not score any other clause,\allowbreak{} and do not let success or failure of another requested edit change the current answer.
\par
The checklist item is the complete success criterion. Answer "yes" only when every explicit qualifier in it is visibly satisfied. Ground the target in the ORIGINAL renders using its shape and stable neighboring geometry,\allowbreak{} then compare the same target role in MODIFIED. Hidden geometry,\allowbreak{} implementation state,\allowbreak{} names,\allowbreak{} metadata,\allowbreak{} and inferred connectivity are not visual evidence. For relative changes,\allowbreak{} compare matching camera directions. For counts and relationships,\allowbreak{} require visibly distinguishable members and direct spatial evidence.
\par
\pnote{User message}
\par
\#\# Full modification query (context only; not visual evidence)\newline
\{full\_\allowbreak{}query\}
\par
\#\# Current independently scored query clause\newline
\{query\_\allowbreak{}clause\}
\par
\#\# Current checklist item\newline
\{checklist\_\allowbreak{}item\}
\par
\{item\_\allowbreak{}guidance\}\newline
Six labeled matched ORIGINAL/\allowbreak{}MODIFIED pairs are attached: front,\allowbreak{} back,\allowbreak{} left,\allowbreak{} right,\allowbreak{} top,\allowbreak{} and bottom. Every pair uses the same camera direction and shared union framing. Inspect all pairs and cite the one with the clearest direct evidence.
\par
Return one JSON object with exactly these fields:\newline
\mbox{}\textasciigrave{}\textasciigrave{}\textasciigrave{}json\newline
\mbox{}\{\newline
\mbox{}~~"answer": "yes",\allowbreak{}\newline
\mbox{}~~"view": "right",\allowbreak{}\newline
\mbox{}~~"evidence": "A concise matched-{}view comparison supporting the answer."\newline
\mbox{}\}\newline
\mbox{}\textasciigrave{}\textasciigrave{}\textasciigrave{}
\par
\textasciigrave{}answer\textasciigrave{} is exactly \textasciigrave{}yes\textasciigrave{} or \textasciigrave{}no\textasciigrave{}. \textasciigrave{}view\textasciigrave{} is exactly one of \textasciigrave{}front\textasciigrave{},\allowbreak{} \textasciigrave{}back\textasciigrave{},\allowbreak{} \textasciigrave{}left\textasciigrave{},\allowbreak{} \textasciigrave{}right\textasciigrave{},\allowbreak{} \textasciigrave{}top\textasciigrave{},\allowbreak{} or \textasciigrave{}bottom\textasciigrave{}. For a negative answer,\allowbreak{} cite the view that most clearly demonstrates the missing or contradictory result. Return only the JSON object.
\par
\pnote{\{item\_guidance\} is empty except for item~0 of G1--G4, where it is:}
\par
This is the framework-{}owned preservation item. Treat every change explicitly requested anywhere in the full query as intended and acceptable,\allowbreak{} even when the change is substantial. Do not require targeted parts to retain their original geometry. Answer "no" only when major non-{}targeted parts are severely corrupted or lost,\allowbreak{} or when the source object becomes unidentifiable or visually incoherent.
\par
\pnote{The text is followed by the twelve renders, each preceded by a text label:}
\par
[ORIGINAL --- front view]\newline
[MODIFIED --- front view]\newline
[...]\newline
[MODIFIED --- bottom view]
\end{promptbox}

\begin{examplebox}[label=ex:armchair, before upper={\footnotesize\raggedright\setlength{\parindent}{0pt}}]{A G4 (Structural) task with verdicts for two editors}
\textbf{Edit prompt} (as given to the editor): Preserve the object's identity and parts outside the requested edits; Raise the seat high enough to create a large open volume beneath it; place four sturdy corner posts between the raised seat and the lower support level; add exactly two long curved rockers running from front to back beneath the corresponding pairs of corner posts; fit a narrower arched back panel between the two rear posts with a broad horizontal gap above the seat; rebuild both arms as slim rails sloping downward from the tall rear posts to the shorter front posts

\smallskip
\textbf{Checklist} with the verdicts for GPT-6.1-sol (Sol) and Gemma 4 31B (Gem.):\par
\setlength{\tabcolsep}{2pt}
\begin{tabular}{@{}r p{0.74\linewidth} c c@{}}
\toprule
 & Item & Sol & Gem. \\
\midrule
0 & Is the modified result still an identifiable and visually coherent version of the source object, without severe corruption of parts not targeted by the requested edits? & \cmark & \cmark \\
1 & Is the seat raised above a large open volume? & \cmark & \cmark \\
2 & Are four sturdy corner posts visible between the raised seat and the lower support level? & \cmark & \cmark \\
3 & Are there exactly two long curved rockers running from front to back beneath the corresponding post pairs? & \cmark & \xmark \\
4 & Is a narrower arched back panel fitted between the rear posts with a broad horizontal gap above the seat? & \cmark & \xmark \\
5 & Are both arms slim rails sloping downward from the tall rear posts to the shorter front posts? & \cmark & \xmark \\
\bottomrule
\end{tabular}
\end{examplebox}

\begin{figure}[t]
    \centering
    
    \setlength{\tabcolsep}{1pt}
    \begin{tabular}{ccc}
    {\scriptsize Original} & {\scriptsize GPT-6.1-sol} & {\scriptsize Gemma 4 31B} \\
    \includegraphics[width=0.315\linewidth]{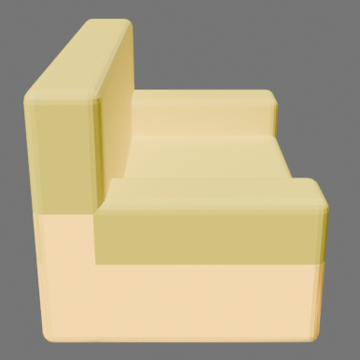} & \includegraphics[width=0.315\linewidth]{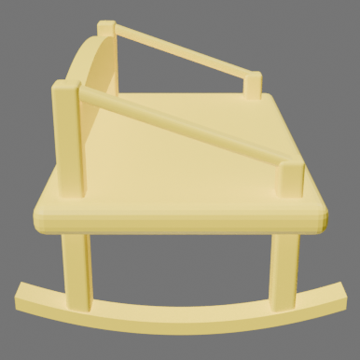} & \includegraphics[width=0.315\linewidth]{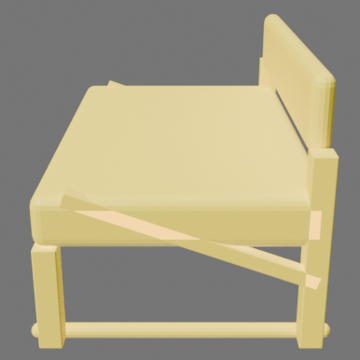} \\
    \includegraphics[width=0.315\linewidth]{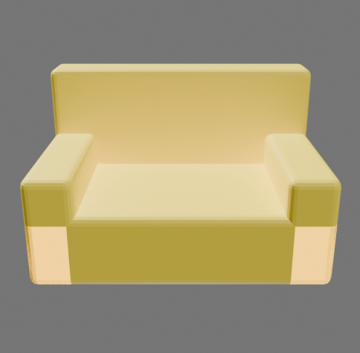} & \includegraphics[width=0.315\linewidth]{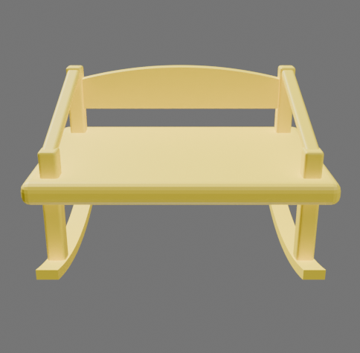} & \includegraphics[width=0.315\linewidth]{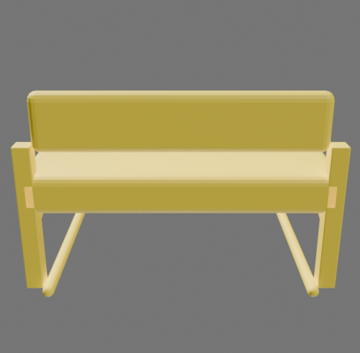} \\
    \end{tabular}
    \caption{\looseness=-1
    Two of the six verifier views for the task of Example~\ref{ex:armchair}.}
    \label{fig:armchair-example}
\end{figure}

\section{Example Tasks and Edits}
\label{app:case-gallery}
\begin{figure*}[tp]
    \centering
    \includegraphics[width=\textwidth]{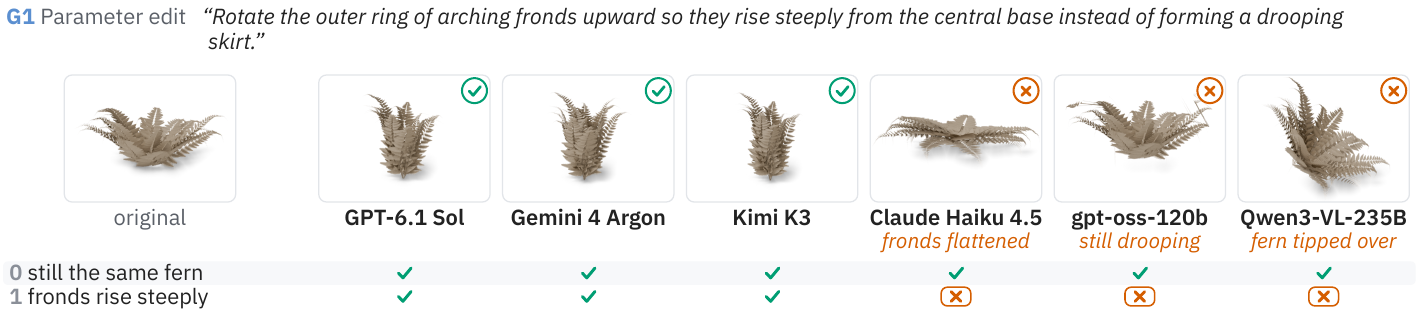}
    \caption{G1: rotate the fern's outer fronds upward so they rise steeply instead of drooping.}
    \label{fig:case-fern}
    \vspace{10pt}
    \includegraphics[width=\textwidth]{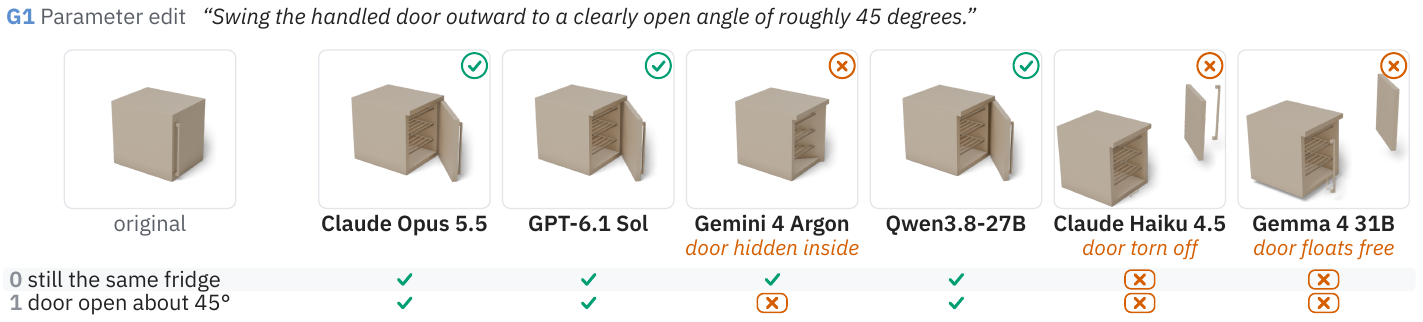}
    \caption{G1: swing the fridge door open to about 45 degrees.}
    \label{fig:case-fridge}
\end{figure*}

\begin{figure*}[tp]
    \centering
    \includegraphics[width=\textwidth]{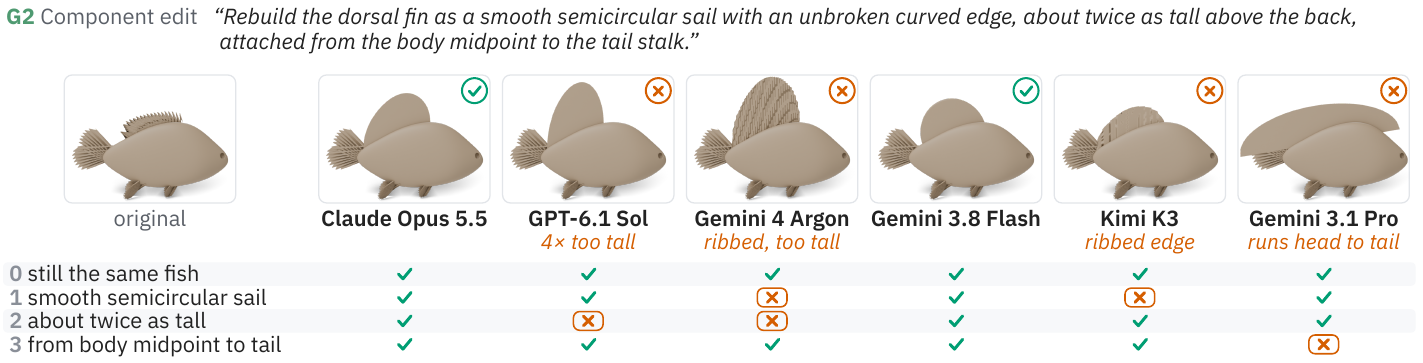}
    \caption{G2: rebuild the dorsal fin as a smooth semicircular sail, twice as tall and longer.}
    \label{fig:case-fish}
    \vspace{10pt}
    \includegraphics[width=\textwidth]{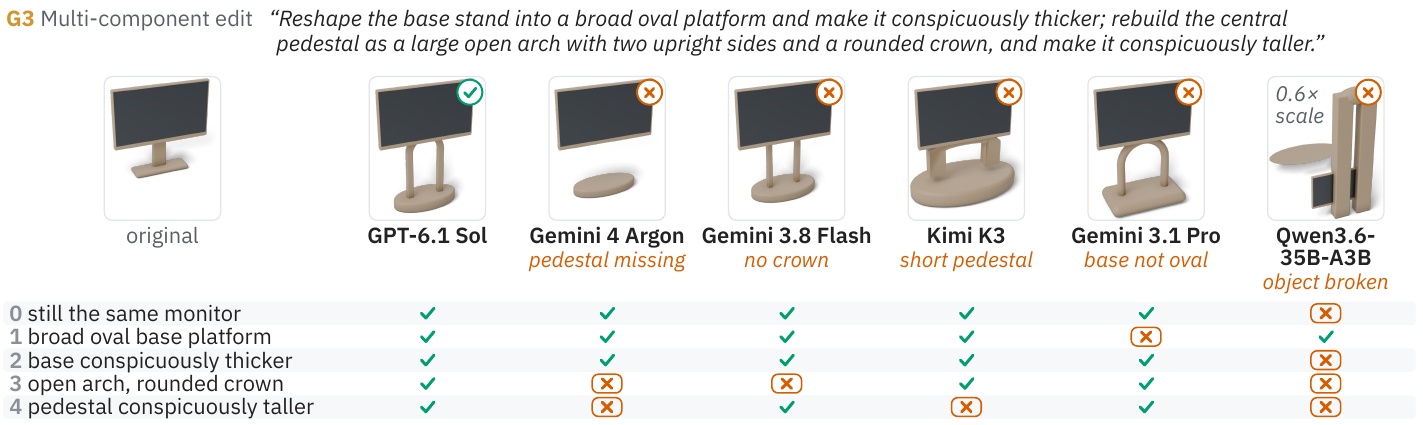}
    \caption{G3: rebuild the monitor stand as a thicker oval base with a taller open arch.}
    \label{fig:case-monitor}
\end{figure*}

\begin{figure*}[tp]
    \centering
    \includegraphics[width=\textwidth]{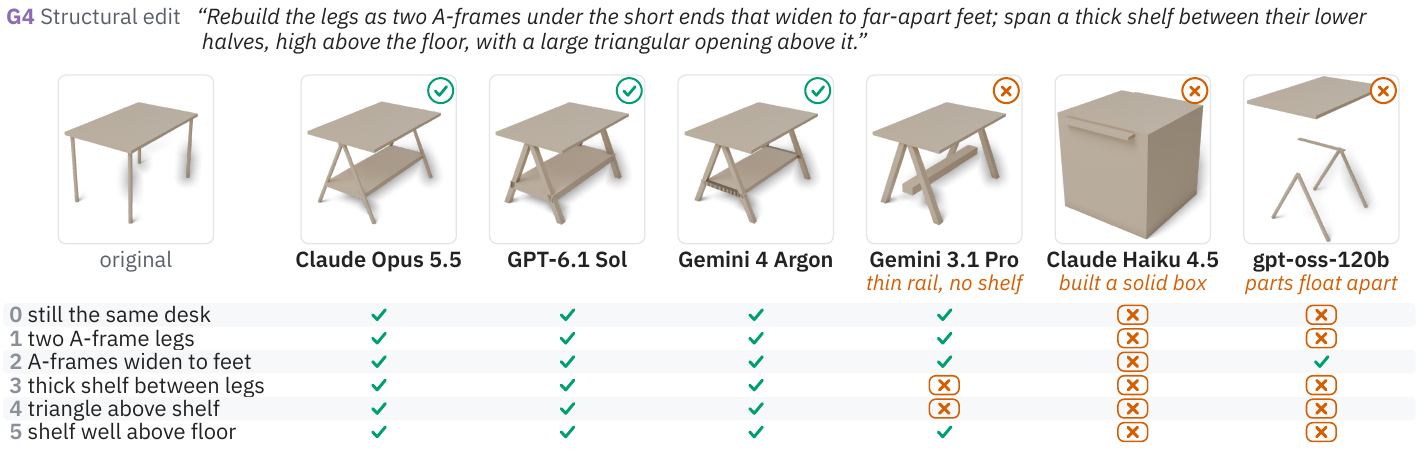}
    \caption{G4: rebuild the desk legs as two A-frames with a thick shelf between them.}
    \label{fig:case-desk}
    \vspace{10pt}
    \includegraphics[width=\textwidth]{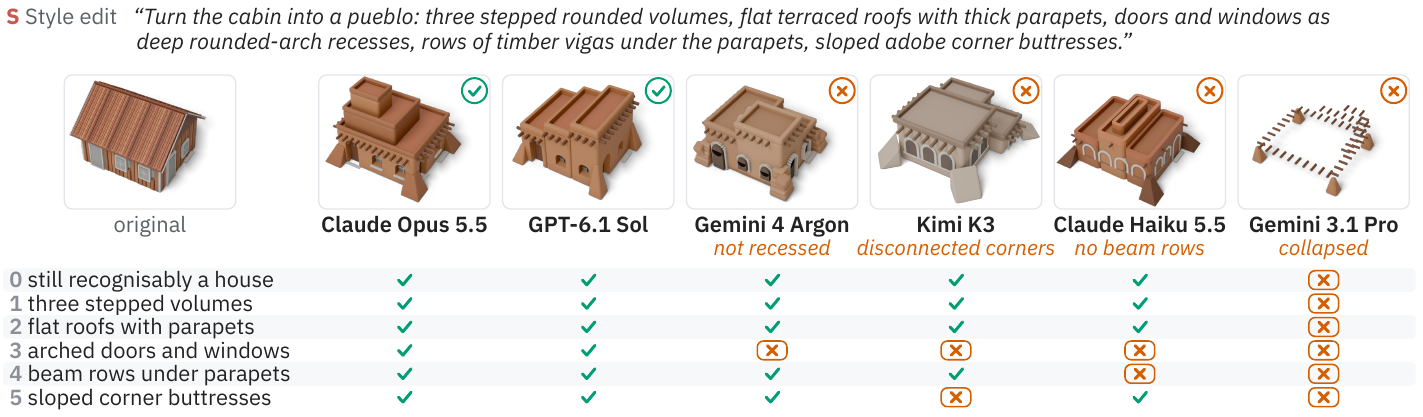}
    \caption{S: turn the log cabin into a stepped adobe pueblo with terraced roofs and timber vigas.}
    \label{fig:case-pueblo}
\end{figure*}

\begin{figure*}[tp]
    \centering
    \includegraphics[width=\textwidth]{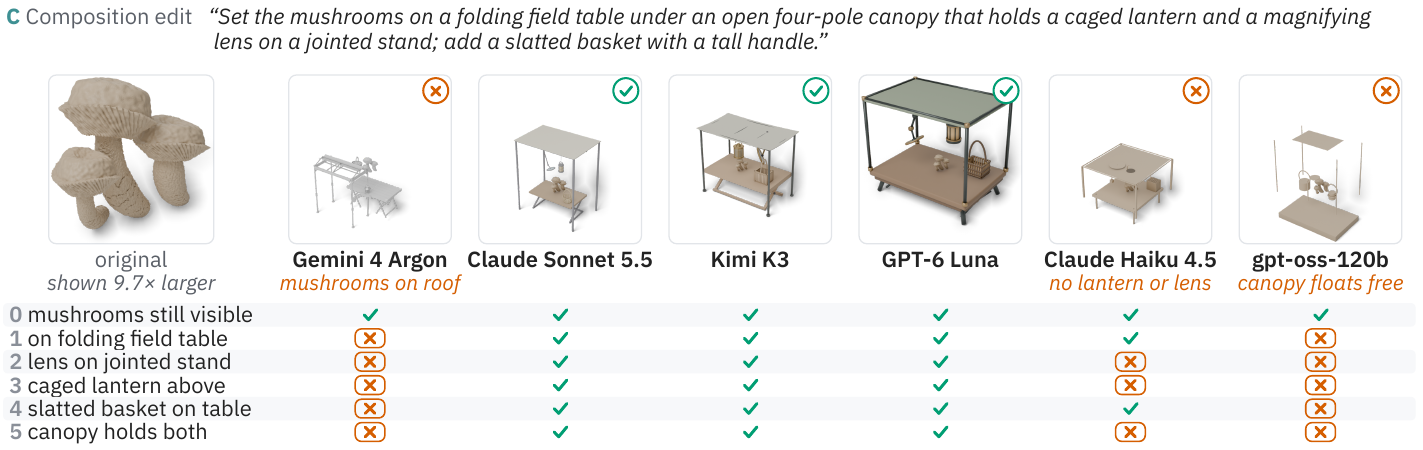}
    \caption{C: place the mushroom cluster on a field table under a canopy holding a lantern and a magnifying lens.}
    \label{fig:case-mushroom}
    \vspace{10pt}
    \includegraphics[width=\textwidth]{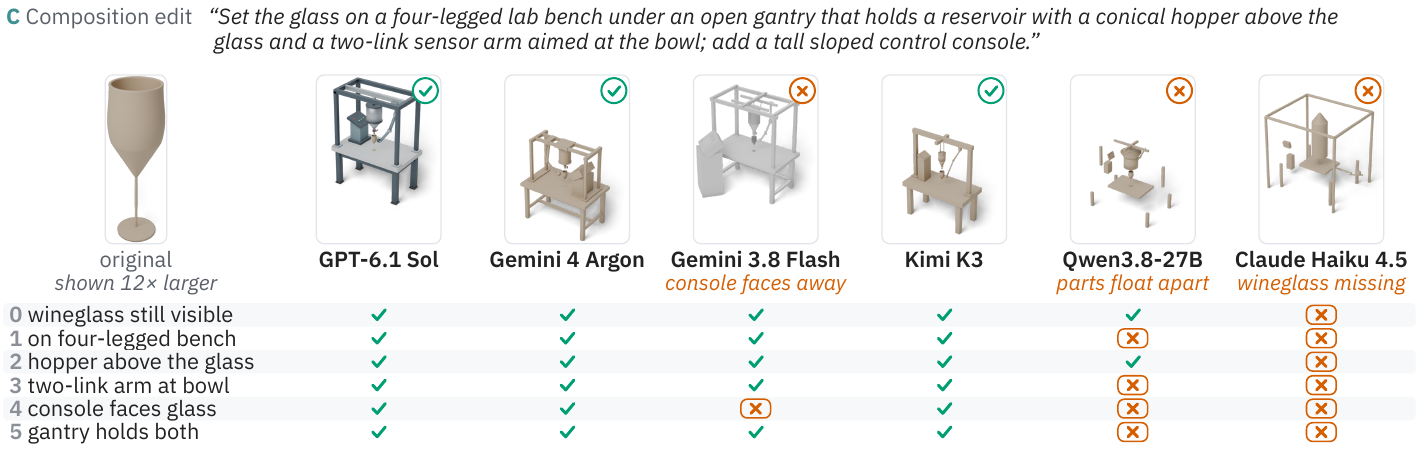}
    \caption{C: place the wineglass on a lab bench under a gantry holding a filling hopper and a sensor arm.}
    \label{fig:case-wineglass}
\end{figure*}

Figures~\ref{fig:case-fern}--\ref{fig:case-wineglass} each show one task: the original object, six edits from the main evaluation by different models, and the checklist judgments for each edit, with a short note on the most visible failure. The cases cover all six edit types, in the order G1, G2, G3, G4, S, and C.

\section{Verifier Reliability}
\label{app:verifier}

\paragraph{Why not standard 3D metrics?}
Standard metrics compare the output either with a reference artifact or with the instruction text, and both fail for open-ended edits. A geometric distance to a single reference penalizes valid edits that realize the instruction differently, while \method{} tasks admit many valid outputs. Image--text similarity is insensitive to the small, localized changes that most edits make, and it cannot verify counts, spatial relations between parts, or whether unrelated parts are preserved. Neither metric separates a correct edit from one that also damages the rest of the artifact.

\paragraph{Human annotation.}
We sample checklist items from the analysis subset across all six edit types, and a human annotator labels each item as pass or fail from the same renders that the verifier sees, yielding 210 items (33--36 per edit type).

\paragraph{Verifier accuracy.}
We compare 16 multimodal models from six providers against the human labels, using the same verifier prompt (Prompt~\ref{prompt:verifier}) and the same renders (Table~\ref{tab:verifier-accuracy}). We report accuracy and Cohen's $\kappa$. Gemini 3.8 Flash with thinking level high, the verifier used throughout, agrees with the human labels on 90.4\% of items ($\kappa=0.80$).

\paragraph{Thinking level and majority voting.}
Spending more computation on each judgment barely changes Flash's agreement. Its accuracy is 87.7\%, 90.0\%, 90.4\%, and 90.8\% at thinking levels low, medium, high, and max. At thinking level high, a majority over three or five calls reaches 91.0\% and 90.5\%, compared with 90.4\% for a single call.

\paragraph{Choice of verifier.}
GPT-6.1-sol agrees more closely with the human labels than any other model (93.8\%, $\kappa=0.87$). We score all experiments with Gemini 3.8 Flash to balance accuracy and cost. The choice barely affects how editors compare: when we score the outputs of nine editors on the analysis subset with both verifiers, the two rankings agree closely (Spearman $\rho=0.98$).

\paragraph{Renders versus code.}
The verifier judges each item from the renders and never reads the program. We test whether the program could replace the renders by giving five verifiers the same 210 items in three forms: the renders, the original and edited programs with their diff, and both (Table~\ref{tab:verifier-modality}). Judging from the programs alone lowers agreement with the human labels for every verifier, by 4.2 to 9.5 points. The reason is that code which reads as correct can still build a different shape. For example, one edit should tilt a rail backward, but its added rotation composes with a later transform in the program and tilts the rail forward. In another, the program defines the requested stepped side rails, yet the rendered object has lost them. Criteria such as ``noticeably wider'' or ``still recognizable'' also depend on how the object looks, which the code does not show directly. In contrast, adding the programs to the renders changes accuracy by only $-0.9$ to $+1.3$ points, so we keep the render-only protocol.

\begin{table}[t]
\centering
\small
\setlength{\tabcolsep}{3pt}
\begin{tabular}{lrrrrrr}
\toprule
& \multicolumn{2}{c}{Renders} & \multicolumn{2}{c}{Code} & \multicolumn{2}{c}{Both} \\
\cmidrule(lr){2-3}\cmidrule(lr){4-5}\cmidrule(lr){6-7}
Verifier & Acc. & $\kappa$ & Acc. & $\kappa$ & Acc. & $\kappa$ \\
\midrule
Gemini 3.8 Flash (high) & 90.4 & 0.80 & 83.5 & 0.62 & 91.7 & 0.82 \\
GPT-6.1-sol & 93.8 & 0.87 & 84.3 & 0.67 & 92.9 & 0.85 \\
GPT-6-astra & 91.9 & 0.83 & 84.3 & 0.67 & 91.9 & 0.83 \\
Claude Opus 5.5 & 91.4 & 0.82 & 85.7 & 0.69 & 91.0 & 0.81 \\
Claude Sonnet 5.5 & 85.2 & 0.69 & 81.0 & 0.57 & 84.3 & 0.67 \\
\bottomrule
\end{tabular}
\caption{Agreement with human labels on the 210 checklist items when the verifier sees the renders (our protocol), the original and edited programs, or both. Acc.\ in \%. Gemini 3.8 Flash averages five calls on renders and three calls otherwise; the other verifiers use one call. GPT-6-astra uses reasoning effort medium here.}
\label{tab:verifier-modality}
\end{table}

\begin{table}[t]
\centering
\small
\setlength{\tabcolsep}{3pt}
\begin{tabular}{lrr}
\toprule
Verifier & Acc. & $\kappa$ \\
\midrule
\multicolumn{3}{l}{\textit{Main verifier and reference}} \\
Gemini 3.8 Flash (low) & 87.7 & 0.74 \\
Gemini 3.8 Flash (medium) & 90.0 & 0.79 \\
Gemini 3.8 Flash (high)$^\ast$ & 90.4 & 0.80 \\
Gemini 3.8 Flash (max) & 90.8 & 0.80 \\
GPT-6.1-sol & \textbf{93.8} & \textbf{0.87} \\
\midrule
\multicolumn{3}{l}{\textit{Other candidates, closed-source}} \\
GPT-5.6-luna & 84.3 & 0.64 \\
GPT-5.6-terra & 83.8 & 0.64 \\
GPT-5.6-sol & 89.5 & 0.77 \\
GPT-6-luna & 82.4 & 0.61 \\
GPT-6-astra & 93.3 & 0.86 \\
Gemini 4 Argon & 93.3 & 0.86 \\
Claude Sonnet 5 & 77.1 & 0.51 \\
Claude Opus 5 & 87.1 & 0.72 \\
Gemini 3.1 Pro & 84.3 & 0.68 \\
Grok 4.6 & 82.9 & 0.62 \\
\midrule
\multicolumn{3}{l}{\textit{Other candidates, open-weight}} \\
Kimi K2.5 & 78.6 & 0.52 \\
Gemma 4 26B-A4B & 71.9 & 0.31 \\
Gemma 4 31B & 81.9 & 0.59 \\
Qwen3-VL-235B-A22B & 71.4 & 0.28 \\
\bottomrule
\end{tabular}
\caption{\looseness=-1
Verifier agreement with human labels on 210 checklist items from the analysis subset. Acc.\ in \%. Single call per item; Gemini 3.8 Flash rows average five independent calls. $^\ast$Configuration used to score all experiments.}
\label{tab:verifier-accuracy}
\end{table}

\section{Multi-Turn Editing Details}
\label{app:multiturn}
All multi-turn experiments use the analysis subset, the editor settings of the main evaluation (reasoning effort medium, 64k output tokens, truncated or refused replies counted as failures), and the main verifier. Syntax and Blender execution errors are returned to the editor as a repair turn.

\paragraph{Prompts and feedback.}
The reviewer works in three steps (Prompt~\ref{prompt:user}). It writes its own checklist from the prompt and the original renders once per task. In each turn, it checks every item in a separate call by splitting the item into qualifiers, and it then writes repair advice for the failed items. The reviewer never sees the benchmark checklist or the verifier's verdicts, and the ``verifier'' in Prompt~\ref{prompt:user}(c) refers to its own item checks. With full feedback, the editor receives a JSON object with the status and evidence of each item and the advice, together with the original and current programs (Prompt~\ref{prompt:revise}). With P/F feedback, it receives only the criterion and status of each item. For text-only editors, self-review reads the original and current programs instead of the renders. In self-check, the editor reviews its own program, and in one setting also its renders, and signals completion by returning the program unchanged. Example~\ref{ex:feedback} is the feedback GPT-6-luna gave itself on the task of Example~\ref{ex:armchair} after its single-shot edit, in which it judged the rockers to run across the chair instead of front to back. In the next turn, the editor reoriented the rockers, and the reviewer accepted the edit. Prompt~\ref{prompt:codex} is the brief of the Codex CLI agent.

\begin{promptbox}[label=prompt:user]{Reviewer with a self-built checklist}
\pnote{(a) Checklist construction, once per task. System prompt:}
\par
Extract a compact,\allowbreak{} complete checklist from the user task and original object. Do not invent constraints or numerical tolerances. Return JSON \{"checklist":[\{"requirement\_\allowbreak{}id":"r1",\allowbreak{}"task\_\allowbreak{}span":"exact quote from task",\allowbreak{}"criterion":"testable requirement"\}]\}. Include task preservation requirements. All three fields are nonempty strings. No other item fields.
\par
\pnote{User message (followed by the six original renders):}
\par
FULL TASK:\newline
\{query\}
\par
\pnote{(b) Item check, one call per checklist item and turn. System prompt:}
\par
You are a skeptical visual inspector of 3D edits. You receive six matched ORIGINAL/\allowbreak{}MODIFIED view pairs (front,\allowbreak{} back,\allowbreak{} left,\allowbreak{} right,\allowbreak{} top,\allowbreak{} bottom),\allowbreak{} the full edit task,\allowbreak{} and ONE requirement. Your job is to find out whether the requirement FAILS.\newline
1. Split the requirement into its explicit qualifiers (which part; count; relative size or ratio; shape; orientation or direction; position; contact or support; what must stay unchanged).\newline
2. For each qualifier,\allowbreak{} choose the view(s) where it is best checked and state exactly what you observe in MODIFIED compared with ORIGINAL. For sizes and ratios,\allowbreak{} compare against a stable reference part in the same view; for counts,\allowbreak{} count visibly separate members; for contact or support,\allowbreak{} look for gaps or floating parts; for orientation,\allowbreak{} state the direction relative to named features.\newline
3. A qualifier passes only if the visible evidence clearly supports it. If it is contradicted or cannot be confirmed,\allowbreak{} it fails. Answer yes only if every qualifier passes. Return only JSON: \{"qualifiers":[\{"qualifier":"...",\allowbreak{}"view":"...",\allowbreak{}"observation":"...",\allowbreak{}"pass":true\}],\allowbreak{}"answer":"yes|no",\allowbreak{}"evidence":"the decisive observation"\}
\par
\pnote{User message (followed by the twelve labeled renders):}
\par
FULL TASK:\newline
\{query\}
\par
REQUIREMENT (quoted task span: \{task\_\allowbreak{}span\}):\newline
\{criterion\}
\par
Six labeled matched ORIGINAL/\allowbreak{}MODIFIED pairs are attached.
\par
\pnote{(c) Repair advice when an item fails. System prompt:}
\par
You diagnose a 3D edit against the complete user task. Use directly visible matched-{}view evidence; do not assume a requested edit succeeded. Do not invent numerical measurements,\allowbreak{} additional requirements or hidden geometry. Inspect preservation as well as intended edits. Distinguish violated from uncertain. Give actionable text feedback,\allowbreak{} not replacement code. Return only the requested JSON object. satisfied means every requirement in the complete task is met,\allowbreak{} with no unresolved uncertainty.
\par
\pnote{User message (followed by the renders); each failed item is listed as ``- \{requirement\_id\} (\{criterion\}): \{evidence\}'':}
\par
FULL TASK:\newline
\{query\}
\par
The following checklist items FAILED independent verification (with the verifier\textquotesingle{}s evidence):\newline
\{failed\}\newline
Items that passed must be preserved. Give concrete,\allowbreak{} actionable repair advice for the failed items only; no replacement code. Return only this JSON object,\allowbreak{} with no other text: \{"feedback":"your repair advice"\}.
\end{promptbox}

\begin{promptbox}[label=prompt:revise]{Editor in multi-turn editing}
\pnote{System prompt}
\par
You are a Blender Python editor. Implement the complete user task,\allowbreak{} preserving non-{}targeted content as required by that task. Treat critic feedback as fallible observations,\allowbreak{} not new requirements. Do not omit helpers or return a patch. Return only one \textasciigrave{}\textasciigrave{}\textasciigrave{}python code block containing the complete executable script,\allowbreak{} with no other text.
\par
\pnote{User message; \{feedback\} is the JSON object returned by the reviewer}
\par
FULL TASK:\newline
\{query\}
\par
ORIGINAL CODE:\newline
\mbox{}\textasciigrave{}\textasciigrave{}\textasciigrave{}python\newline
\mbox{}\{original\_\allowbreak{}code\}\newline
\mbox{}\textasciigrave{}\textasciigrave{}\textasciigrave{}
\par
CURRENT CODE:\newline
\mbox{}\textasciigrave{}\textasciigrave{}\textasciigrave{}python\newline
\mbox{}\{current\_\allowbreak{}code\}\newline
\mbox{}\textasciigrave{}\textasciigrave{}\textasciigrave{}
\par
LATEST CRITIC FEEDBACK:\newline
\{feedback\}
\par
Return only one \textasciigrave{}\textasciigrave{}\textasciigrave{}python code block containing the complete executable script,\allowbreak{} with no other text.
\par
\pnote{After a syntax or Blender execution error, the feedback block is replaced by:}
\par
CURRENT EXECUTION FAILURE (repair it):\newline
\{error\}
\par
Review the current code and any supplied images yourself. Correct failures; if already satisfied,\allowbreak{} preserve the code.
\par
\pnote{In self-check, it is replaced by:}
\par
Review the current code and any supplied images yourself. If the current code already satisfies the complete task,\allowbreak{} return it strictly unchanged to signal that you are satisfied. Otherwise,\allowbreak{} return the complete revised script.
\end{promptbox}

\begin{examplebox}[label=ex:feedback]{Feedback from the reviewer (abbreviated)}
\mbox{}\{"status": "needs\_\allowbreak{}revision",\allowbreak{}\newline
\mbox{}~"checks": [\newline
\mbox{}~~\{"requirement\_\allowbreak{}id": "r1",\allowbreak{} "status": "satisfied",\allowbreak{}\newline
\mbox{}~~~"evidence": "The modified object remains recognizably the same yellow chair,\allowbreak{} [...]"\},\allowbreak{}\newline
\mbox{}~~[...]\newline
\mbox{}~~\{"requirement\_\allowbreak{}id": "r4",\allowbreak{} "status": "violated",\allowbreak{}\newline
\mbox{}~~~"evidence": "The two curved bars are oriented across the chair\textquotesingle{}s width in the bottom view,\allowbreak{} not front to back. Failing qualifiers: [...]"\},\allowbreak{}\newline
\mbox{}~~[...]],\allowbreak{}\newline
\mbox{}~"feedback": "Reorient the two curved bottom bars so each runs lengthwise from front to back beneath one side of the chair,\allowbreak{} connecting the front and rear corner posts on that side. The current bars span left-{}to-{}right between the posts at each end; remove those transverse bars and retain exactly two curved rockers,\allowbreak{} one under each front-{}to-{}back post pair."\}
\end{examplebox}

\begin{promptbox}[label=prompt:codex]{Codex CLI agent brief (abbreviated)}
You are an expert Blender Python programmer working in the current directory.
\par
\#\# Task\newline
Modify the Blender Python script \textasciigrave{}original.py\textasciigrave{} according to the modification instruction in \textasciigrave{}task.md\textasciigrave{},\allowbreak{} and save the complete modified script as \textasciigrave{}edited.py\textasciigrave{} in this directory (\textasciigrave{}edited.py\textasciigrave{} currently contains an unmodified copy of \textasciigrave{}original.py\textasciigrave{}).\newline
-{} Implement every requested clause of the instruction.\newline
-{} Preserve unrelated behavior,\allowbreak{} definitions,\allowbreak{} imports,\allowbreak{} helpers and geometry. Make no additional visible edits beyond those required by the instruction.\newline
-{} \textasciigrave{}edited.py\textasciigrave{} must be a complete,\allowbreak{} executable script (not a patch or diff).\newline
\par
\#\# Rendering tool\newline
\textasciigrave{}./\allowbreak{}render.sh edited.py\textasciigrave{} runs the script headless in Blender 5.1 and renders six matched views (front,\allowbreak{} back,\allowbreak{} left,\allowbreak{} right,\allowbreak{} top,\allowbreak{} bottom) of the ORIGINAL object and of the MODIFIED result,\allowbreak{} from identical cameras. It prints whether the script ran and the image paths: \textasciigrave{}renders/\allowbreak{}rNN/\allowbreak{}sheet.png\textasciigrave{} (top row ORIGINAL,\allowbreak{} bottom row MODIFIED,\allowbreak{} columns front/\allowbreak{}back/\allowbreak{}left/\allowbreak{}right/\allowbreak{}top/\allowbreak{}bottom) and one PNG per view and state. If the script fails,\allowbreak{} it prints the error. Open the images to inspect your result.\newline
-{} You have a budget of \{budget\} render calls in total. Failed renders count. After the budget is used,\allowbreak{} \textasciigrave{}render.sh\textasciigrave{} refuses further calls.
\par
[...]
\par
\#\# Rules\newline
-{} Work only with files in this directory. Do not read,\allowbreak{} list or search files outside it.\newline
-{} Do not run Blender or any other renderer directly; rendering is only available through \textasciigrave{}./\allowbreak{}render.sh\textasciigrave{}.\newline
-{} There is no internet access. The session has a wall-{}clock limit of \{wall\_\allowbreak{}min\} minutes; whatever \textasciigrave{}edited.py\textasciigrave{} contains at the end is your answer.
\par
When you are finished,\allowbreak{} make sure \textasciigrave{}edited.py\textasciigrave{} contains your final script,\allowbreak{} then reply with one line: DONE
\par
\pnote{Variant with self-review: after the agent finishes, the reviewer of Prompt~\ref{prompt:user} inspects the renders, and the session is resumed with}
\par
An independent reviewer inspected the rendered result of your current \textasciigrave{}edited.py\textasciigrave{} against the instruction and found that it does NOT yet satisfy it.
\par
Reviewer findings (requirements that failed,\allowbreak{} with the visual evidence):\newline
\{failed\}
\par
Reviewer advice:\newline
\{advice\}
\par
Treat these findings as fallible observations,\allowbreak{} not new requirements. Revise \textasciigrave{}edited.py\textasciigrave{} to satisfy the complete instruction in \textasciigrave{}task.md\textasciigrave{} (you may use \textasciigrave{}./\allowbreak{}render.sh\textasciigrave{} within the remaining budget),\allowbreak{} then reply with one line: DONE
\end{promptbox}

\looseness=-1
\paragraph{Revision rounds.}
Table~\ref{tab:mt-rounds} gives the mean number of revisions in each setting of Table~\ref{tab:mt-main}. When checking themselves, strong editors rarely revise, whereas a reviewer keeps them revising for several rounds.

\begin{table}[t]
\centering
\footnotesize
\setlength{\tabcolsep}{3pt}
\begin{tabular}{@{}lrrrrr@{}}
\toprule
 & \multicolumn{2}{c}{Self-check} & \multicolumn{2}{c}{Indep.\ rev.} & \\
\cmidrule(lr){2-3}\cmidrule(lr){4-5}
Editor & Code & Render & P/F & Full & Self-rev. \\
\midrule
GPT-6.1-sol & 0.3 & 0.3 & 3.4 & 2.8 & 1.5 \\
Claude Opus 5.5 & 0.2 & 0.5 & 3.5 & 2.7 & 1.6 \\
Claude Sonnet 5.5 & 0.4 & 0.6 & 4.3 & 3.8 & 4.7 \\
GPT-6-luna & 1.8 & 2.7 & 4.5 & 3.2 & 4.2 \\
Qwen3.8-27B & 1.5 & 1.6 & 5.5 & 5.0 & 2.6 \\
Qwen3-VL-235B & 2.1 & 2.1 & 7.0 & 6.9 & 4.4 \\
gpt-oss-120b & 1.6 & -- & 6.7 & 6.5 & 3.1 \\
Qwen3-Coder-Next & 4.6 & -- & 7.1 & 7.0 & 4.3 \\
\bottomrule
\end{tabular}
\caption{Mean rounds of revision in each multi-turn setting of Table~\ref{tab:mt-main} (analysis subset, at most 8). --: text-only editor.}
\label{tab:mt-rounds}
\end{table}

\paragraph{Coding agent harness.}
\label{app:codex}
We also run Codex CLI with GPT-6-luna and GPT-6.1-sol at reasoning effort medium, a tool that renders the same six views as the verifier (at most 9 renders), and a 90-minute limit per turn. In the plain harness, the agent decides when it is done. With self-review, the editor's own model reviews the renders after the agent finishes, as in \S\ref{sec:exp-multiturn}, and the session resumes with its feedback for up to 8 rounds. The plain harness helps GPT-6.1-sol (64.7\% to 74.7\%) but not GPT-6-luna (36.9\% to 34.6\%), and stays below multi-turn editing. With the reviewer providing feedback and deciding when the edit is done, the same harness matches multi-turn editing (Table~\ref{tab:mt-agent}).

\begin{table}[t]
\centering
\footnotesize
\setlength{\tabcolsep}{4pt}
\begin{tabular}{@{}lrr@{}}
\toprule
Setting & Luna & Sol \\
\midrule
Single-shot edit & 36.9 & 64.7 \\
Single-shot edit, best effort & 54.2 & 74.7 \\
Codex harness & 34.6 & 74.7 \\
Codex harness + self-review & 59.6 & 84.9 \\
Multi-turn editing (self-review) & 61.5 & 83.0 \\
\bottomrule
\end{tabular}
\caption{All-pass rate (\%) of a coding agent (Codex CLI) and multi-turn editing on the analysis subset. Luna: GPT-6-luna; Sol: GPT-6.1-sol. Best effort: max for Luna, xhigh for Sol.}
\label{tab:mt-agent}
\end{table}

\renewcommand{\topfraction}{0.85}\renewcommand{\dbltopfraction}{0.85}\renewcommand{\textfraction}{0.1}
\renewcommand{\floatpagefraction}{0.5}\renewcommand{\dblfloatpagefraction}{0.5}

\section{Additional Results}
\label{app:results}
\begin{table*}[t]
\centering
\small
\setlength{\tabcolsep}{6pt}
\begin{tabular}{lrrrrrrrr}
\toprule
 & \multicolumn{7}{c}{All-pass rate} & \\
\cmidrule(lr){2-8}
Model & G1 & G2 & G3 & G4 & S & C & All & Exec. \\
\midrule
Claude Opus 5.5 & \textbf{96.5} & \textbf{76.9} & 64.4 & 58.6 & 52.5 & \textbf{70.8} & \textbf{69.9} & \textbf{99.3} \\
GPT-6.1-sol & 95.6 & 71.3 & \textbf{65.3} & \textbf{59.3} & \textbf{57.6} & 65.7 & 69.1 & 96.1 \\
Gemini 4 Argon & 94.7 & 67.4 & 54.2 & 41.0 & 41.0 & 53.0 & 58.5 & 95.4 \\
Gemini 3.8 Flash & 94.7 & 68.8 & 62.7 & 53.7 & 34.5 & 36.3 & 58.4 & 92.0 \\
Claude Sonnet 5.5 & 91.4 & 66.4 & 51.2 & 41.9 & 25.9 & 45.4 & 53.7 & 98.8 \\
Kimi K3 & 88.2 & 57.9 & 47.5 & 38.9 & 27.3 & 42.4 & 50.3 & 96.5 \\
Claude Haiku 5.5 & 90.3 & 54.4 & 42.8 & 31.5 & 16.0 & 32.9 & 44.6 & 94.3 \\
GPT-6-luna & 87.0 & 51.4 & 36.6 & 29.4 & 18.8 & 26.9 & 41.7 & 92.4 \\
Gemini 3.1 Pro & 88.7 & 50.9 & 41.0 & 28.0 & 7.9 & 8.3 & 37.5 & 96.4 \\
Qwen3.8-27B & 84.3 & 44.2 & 28.5 & 17.1 & 3.5 & 1.6 & 29.9 & 89.3 \\
Qwen3.6-27B & 80.8 & 29.6 & 16.0 & 6.0 & 0.2 & 0.0 & 22.1 & 81.8 \\
Claude Haiku 4.5 & 81.5 & 28.9 & 9.3 & 4.4 & 0.0 & 0.0 & 20.7 & 91.6 \\
Gemma 4 31B & 77.5 & 21.5 & 11.8 & 4.9 & 0.2 & 0.0 & 19.3 & 86.2 \\
Qwen3.5-27B & 78.5 & 20.6 & 8.1 & 1.9 & 0.0 & 0.0 & 18.2 & 68.6 \\
Qwen3.6-35B-A3B & 70.6 & 18.8 & 5.3 & 1.2 & 0.0 & 0.0 & 16.0 & 72.7 \\
gpt-oss-120b & 71.5 & 18.1 & 4.4 & 1.6 & 0.0 & 0.0 & 15.9 & 77.3 \\
Qwen3.5-35B-A3B & 68.3 & 7.6 & 2.8 & 0.0 & 0.0 & 0.0 & 13.1 & 63.9 \\
Qwen3-VL-235B-A22B & 61.8 & 9.0 & 2.3 & 0.5 & 0.0 & 0.0 & 12.3 & 68.4 \\
Qwen3-Coder-Next & 54.2 & 4.9 & 1.4 & 0.2 & 0.0 & 0.0 & 10.1 & 51.1 \\
Qwen3.5-9B & 55.3 & 3.0 & 1.4 & 0.2 & 0.0 & 0.0 & 10.0 & 47.1 \\
Qwen3.5-4B & 48.8 & 0.9 & 0.0 & 0.0 & 0.0 & 0.0 & 8.3 & 38.8 \\
Qwen3.5-2B & 2.1 & 0.0 & 0.0 & 0.0 & 0.0 & 0.0 & 0.3 & 15.8 \\
\midrule
Average & 75.6 & 35.1 & 25.3 & 19.1 & 13.0 & 17.4 & 30.9 & 77.9 \\
\bottomrule
\end{tabular}
\caption{All-pass rate (\%) of all 22 evaluated models by edit type (G1--G4: geometric edits graded by scope; S: style; C: composition), sorted by overall rate; the best value in each column is in bold. Exec.: percentage of edited programs that execute and render. Average: mean over the 22 models.}
\label{tab:main-results-full}
\end{table*}

\begin{table*}[p]
\centering
\small
\setlength{\tabcolsep}{5pt}
\begin{tabular}{llrrrrrrrrr}
\toprule
Model & Effort & G1 & G2 & G3 & G4 & S & C & All & Exec. & Tokens \\
\midrule
Claude Opus 5.5 & low & 94.2 & 61.5 & 48.1 & 51.9 & 50.0 & 53.8 & 59.9 & 100.0 & 11.1 \\
 & medium & 96.2 & 73.1 & 55.8 & 51.9 & 51.9 & 76.9 & 67.6 & 99.4 & 14.7 \\
 & high & 98.1 & 71.2 & 61.5 & 57.7 & 51.9 & 73.1 & 68.9 & 99.4 & 17.9 \\
 & xhigh & 94.2 & 71.2 & 69.2 & 57.7 & 61.5 & 78.8 & \textbf{72.1} & 99.7 & 38.8 \\
\midrule
GPT-6.1-sol & low & 94.2 & 73.1 & 59.6 & 59.6 & 51.9 & 63.5 & 67.0 & 96.2 & 7.1 \\
 & medium & 96.2 & 63.5 & 53.8 & 51.9 & 48.1 & 75.0 & 64.7 & 94.6 & 7.8 \\
 & high & 98.1 & 69.2 & 61.5 & 63.5 & 55.8 & 76.9 & 70.8 & 93.6 & 9.9 \\
 & xhigh & 96.2 & 75.0 & 69.2 & 69.2 & 67.3 & 71.2 & \textbf{74.7} & 94.6 & 13.6 \\
 & max & 94.2 & 71.2 & 55.8 & 71.2 & 61.5 & 69.2 & 70.5 & 93.3 & 25.6 \\
\midrule
Gemini 3.8 Flash & low & 92.3 & 50.0 & 42.3 & 32.7 & 7.7 & 21.2 & 41.0 & 90.7 & 8.6 \\
 & medium & 94.2 & 67.3 & 55.8 & 63.5 & 25.0 & 40.4 & 57.7 & 90.1 & 29.6 \\
 & high & 86.5 & 48.1 & 44.2 & 48.1 & 40.4 & 34.6 & 50.3 & 80.1 & 45.8 \\
 & max & 92.3 & 71.2 & 63.5 & 57.7 & 30.8 & 36.5 & \textbf{58.7} & 93.3 & 29.5 \\
\midrule
Claude Sonnet 5.5 & low & 94.2 & 50.0 & 30.8 & 28.8 & 23.1 & 30.8 & 42.9 & 98.4 & 9.8 \\
 & medium & 88.5 & 57.7 & 42.3 & 40.4 & 28.8 & 46.2 & 50.6 & 98.7 & 11.7 \\
 & high & 96.2 & 61.5 & 55.8 & 63.5 & 38.5 & 76.9 & 65.4 & 99.0 & 18.5 \\
 & xhigh & 94.2 & 71.2 & 67.3 & 71.2 & 51.9 & 86.5 & \textbf{73.7} & 97.8 & 41.9 \\
\midrule
Kimi K3 & none & 78.8 & 30.8 & 11.5 & 13.5 & 5.8 & 5.8 & 24.4 & 93.9 & 8.3 \\
 & low & 86.5 & 36.5 & 17.3 & 17.3 & 11.5 & 15.4 & 30.8 & 93.9 & 7.7 \\
 & medium & 86.5 & 57.7 & 38.5 & 28.8 & 26.9 & 42.3 & 46.8 & 94.2 & 14.5 \\
 & high & 90.4 & 67.3 & 44.2 & 36.5 & 19.2 & 44.2 & 50.3 & 96.2 & 14.6 \\
 & xhigh & 92.3 & 63.5 & 46.2 & 40.4 & 30.8 & 50.0 & 53.8 & 96.8 & 26.5 \\
 & max & 90.4 & 65.4 & 44.2 & 48.1 & 30.8 & 48.1 & \textbf{54.5} & 96.2 & 25.4 \\
\midrule
GPT-6-luna & low & 86.5 & 40.4 & 19.2 & 13.5 & 3.8 & 3.8 & 27.9 & 88.8 & 6.3 \\
 & medium & 90.4 & 42.3 & 28.8 & 17.3 & 15.4 & 26.9 & 36.9 & 90.7 & 9.2 \\
 & high & 84.6 & 59.6 & 32.7 & 34.6 & 13.5 & 32.7 & 42.9 & 89.7 & 14.2 \\
 & xhigh & 92.3 & 59.6 & 42.3 & 42.3 & 34.6 & 38.5 & 51.6 & 93.3 & 19.1 \\
 & max & 92.3 & 53.8 & 48.1 & 44.2 & 38.5 & 48.1 & \textbf{54.2} & 97.4 & 41.7 \\
\midrule
Claude Sonnet 4.6 & low & 82.7 & 40.4 & 26.9 & 21.2 & 3.8 & 1.9 & 29.5 & 98.4 & 17.6 \\
 & medium & 88.5 & 48.1 & 32.7 & 25.0 & 3.8 & 19.2 & \textbf{36.2} & 94.9 & 41.7 \\
 & high & 86.5 & 46.2 & 30.8 & 28.8 & 3.8 & 19.2 & 35.9 & 84.9 & 69.3 \\
\midrule
Claude Sonnet 5 & low & 78.8 & 25.0 & 15.4 & 19.2 & 3.8 & 1.9 & 24.0 & 96.8 & 12.6 \\
 & medium & 80.8 & 44.2 & 23.1 & 25.0 & 0.0 & 0.0 & 28.8 & 96.2 & 17.2 \\
 & high & 86.5 & 48.1 & 26.9 & 25.0 & 5.8 & 19.2 & 35.3 & 96.8 & 26.6 \\
 & xhigh & 92.3 & 51.9 & 34.6 & 23.1 & 13.5 & 25.0 & \textbf{40.1} & 97.4 & 42.8 \\
\midrule
GPT-5.6-luna & none & 67.3 & 17.3 & 5.8 & 1.9 & 3.8 & 0.0 & 16.0 & 78.2 & 5.3 \\
 & low & 53.8 & 28.8 & 9.6 & 17.3 & 0.0 & 0.0 & 18.3 & 65.7 & 3.1 \\
 & medium & 69.2 & 30.8 & 15.4 & 19.2 & 17.3 & 5.8 & 26.3 & 73.1 & 5.0 \\
 & high & 90.4 & 51.9 & 36.5 & 30.8 & 21.2 & 23.1 & 42.3 & 92.0 & 13.1 \\
 & xhigh & 92.3 & 53.8 & 42.3 & 32.7 & 13.5 & 38.5 & 45.5 & 94.2 & 20.0 \\
 & max & 94.2 & 63.5 & 48.1 & 46.2 & 30.8 & 51.9 & \textbf{55.8} & 96.5 & 33.7 \\
\bottomrule
\end{tabular}
\caption{All-pass rate (\%) by reasoning effort on the analysis subset (312 tasks; 52 per edit type). For the six models that are also in Table~\ref{tab:main-results-full}, the medium level is their main run (at most 64k output tokens). All other levels use the provider's maximum output length: 128k tokens for Claude and Kimi K3, 65,536 for Gemini 3.8 Flash, and no limit for GPT models. At this limit, the Claude models at max effort are truncated on up to 54\% of a 24-task pilot, so we omit that level. Exec.: percentage of edited programs that execute and render. Tokens: mean output tokens per task in thousands, including reasoning tokens. The best overall rate of each model is in bold.}
\label{tab:effort-sweep}
\end{table*}

Table~\ref{tab:main-results-full} reports the all-pass rate of all 22 evaluated models, including the five Qwen3.5 models omitted from Table~\ref{tab:main-results}.

\paragraph{Reasoning effort.}
Table~\ref{tab:effort-sweep} reports every effort level of the nine models whose effort we vary on the analysis subset. The most useful level differs by model: GPT-6.1-sol peaks at xhigh (74.7\%) and declines at max (70.5\%) despite nearly twice the output tokens, whereas Gemini 3.8 Flash, GPT-6-luna, GPT-5.6-luna, and Kimi K3 are best at max. The gains concentrate on the harder types, for example 5.7 points on G1 against 23 to 40 points on G3, G4, S, and C for Claude Sonnet 5.5 from medium to xhigh.

\begin{table*}[p]
\centering
\small
\setlength{\tabcolsep}{4.5pt}
\begin{tabular}{lllrrrrrrrr}
\toprule
Model & Released & Family & G1 & G2 & G3 & G4 & S & C & All & Exec. \\
\midrule
Qwen2.5-Coder-32B & 2024-11-12 & Qwen 27--32B & 50.0 & 3.8 & 0.0 & 0.0 & 0.0 & 0.0 & 9.0 & 57.7 \\
QwQ-32B & 2025-03-06 & Qwen 27--32B & 46.2 & 5.8 & 0.0 & 1.9 & 0.0 & 0.0 & 9.0 & 41.3 \\
Qwen3-32B & 2025-04-29 & Qwen 27--32B & 48.1 & 9.6 & 0.0 & 0.0 & 0.0 & 0.0 & 9.6 & 52.2 \\
Claude Sonnet 4 & 2025-05-22 & Claude Sonnet & 71.2 & 21.2 & 7.7 & 3.8 & 0.0 & 0.0 & 17.3 & 89.4 \\
Gemini 2.5 Flash & 2025-06-17 & Gemini Flash & 67.3 & 13.5 & 7.7 & 5.8 & 0.0 & 0.0 & 15.7 & 56.7 \\
gpt-oss-120b\smash{$^\ast$} & 2025-08-05 & gpt-oss & 75.0 & 9.6 & 0.0 & 1.9 & 0.0 & 0.0 & 14.4 & 75.3 \\
Qwen3-VL-235B-A22B\smash{$^\ast$} & 2025-09-23 & Qwen other & 61.5 & 7.7 & 1.9 & 0.0 & 0.0 & 0.0 & 11.9 & 72.8 \\
Claude Sonnet 4.5 & 2025-09-29 & Claude Sonnet & 76.9 & 30.8 & 5.8 & 7.7 & 1.9 & 0.0 & 20.5 & 92.0 \\
Claude Haiku 4.5\smash{$^\ast$} & 2025-10-15 & Claude Haiku & 80.8 & 19.2 & 3.8 & 7.7 & 0.0 & 0.0 & 18.6 & 90.1 \\
Claude Opus 4.5 & 2025-11-24 & Claude Opus & 88.5 & 51.9 & 13.5 & 23.1 & 3.8 & 1.9 & 30.4 & 95.8 \\
Gemini 3 Flash & 2025-12-17 & Gemini Flash & 82.7 & 44.2 & 25.0 & 13.5 & 1.9 & 0.0 & 27.9 & 88.1 \\
Qwen3-Coder-Next\smash{$^\ast$} & 2026-02-01\smash{$^\dagger$} & Qwen other & 51.9 & 0.0 & 0.0 & 0.0 & 0.0 & 0.0 & 8.7 & 45.2 \\
Claude Opus 4.6 & 2026-02-05 & Claude Opus & 88.5 & 46.2 & 36.5 & 26.9 & 7.7 & 0.0 & 34.3 & 95.8 \\
Claude Sonnet 4.6 & 2026-02-17 & Claude Sonnet & 88.5 & 48.1 & 32.7 & 25.0 & 3.8 & 19.2 & 36.2 & 94.9 \\
Gemini 3.1 Pro\smash{$^\ast$} & 2026-02-19 & Gemini Pro & 88.5 & 50.0 & 32.7 & 26.9 & 5.8 & 7.7 & 35.3 & 96.5 \\
Qwen3.5-27B\smash{$^\ast$} & 2026-02-24\smash{$^\dagger$} & Qwen 27--32B & 78.8 & 15.4 & 0.0 & 1.9 & 0.0 & 0.0 & 16.0 & 67.9 \\
Qwen3.5-35B-A3B\smash{$^\ast$} & 2026-02-24\smash{$^\dagger$} & Qwen other & 61.5 & 3.8 & 3.8 & 0.0 & 0.0 & 0.0 & 11.5 & 68.9 \\
Qwen3.5-2B\smash{$^\ast$} & 2026-03-02\smash{$^\dagger$} & Qwen other & 1.9 & 0.0 & 0.0 & 0.0 & 0.0 & 0.0 & 0.3 & 7.4 \\
Qwen3.5-4B\smash{$^\ast$} & 2026-03-02\smash{$^\dagger$} & Qwen other & 55.8 & 1.9 & 0.0 & 0.0 & 0.0 & 0.0 & 9.6 & 38.5 \\
Qwen3.5-9B\smash{$^\ast$} & 2026-03-02\smash{$^\dagger$} & Qwen other & 46.2 & 0.0 & 0.0 & 0.0 & 0.0 & 0.0 & 7.7 & 45.8 \\
GPT-5.4 & 2026-03-05 & GPT flagship & 94.2 & 44.2 & 32.7 & 23.1 & 9.6 & 36.5 & 40.1 & 96.8 \\
Gemma 4 31B\smash{$^\ast$} & 2026-04-01\smash{$^\dagger$} & Gemma & 75.0 & 13.5 & 1.9 & 5.8 & 0.0 & 0.0 & 16.0 & 83.7 \\
Qwen3.6-35B-A3B\smash{$^\ast$} & 2026-04-16 & Qwen other & 67.3 & 11.5 & 0.0 & 1.9 & 0.0 & 0.0 & 13.5 & 67.6 \\
Qwen3.6-27B\smash{$^\ast$} & 2026-04-22 & Qwen 27--32B & 82.7 & 15.4 & 5.8 & 13.5 & 0.0 & 0.0 & 19.6 & 79.5 \\
GPT-5.5 & 2026-04-23 & GPT flagship & 82.7 & 50.0 & 48.1 & 44.2 & 42.3 & 42.3 & 51.6 & 96.2 \\
Gemini 3.5 Flash & 2026-05-19 & Gemini Flash & 90.4 & 48.1 & 34.6 & 30.8 & 21.2 & 7.7 & 38.8 & 89.7 \\
Claude Opus 4.8 & 2026-05-28 & Claude Opus & 84.6 & 51.9 & 34.6 & 30.8 & 15.4 & 19.2 & 39.4 & 97.4 \\
Claude Sonnet 5 & 2026-06-30 & Claude Sonnet & 80.8 & 44.2 & 23.1 & 25.0 & 0.0 & 0.0 & 28.8 & 96.2 \\
GPT-5.6-luna & 2026-07-09 & GPT small & 69.2 & 30.8 & 15.4 & 19.2 & 17.3 & 5.8 & 26.3 & 73.1 \\
GPT-5.6-sol & 2026-07-09 & GPT flagship & 94.2 & 55.8 & 51.9 & 44.2 & 36.5 & 36.5 & 53.2 & 95.2 \\
Kimi K3\smash{$^\ast$} & 2026-07-16 & Kimi & 86.5 & 57.7 & 38.5 & 28.8 & 26.9 & 42.3 & 46.8 & 94.2 \\
Gemini 3.6 Flash & 2026-07-21 & Gemini Flash & 92.3 & 59.6 & 34.6 & 26.9 & 15.4 & 19.2 & 41.3 & 92.0 \\
Claude Opus 5 & 2026-07-24 & Claude Opus & 92.3 & 61.5 & 46.2 & 40.4 & 32.7 & 40.4 & 52.2 & 98.4 \\
Gemini 3.7 Flash & 2026-08-13 & Gemini Flash & 96.2 & 63.5 & 40.4 & 36.5 & 17.3 & 17.3 & 45.2 & 90.4 \\
Qwen3.8-27B\smash{$^\ast$} & 2026-08-14 & Qwen 27--32B & 76.9 & 38.5 & 15.4 & 21.2 & 3.8 & 0.0 & 26.0 & 90.1 \\
Claude Fable 5.1 & 2026-09-01 & Claude Fable & 92.3 & 57.7 & 48.1 & 42.3 & 26.9 & 55.8 & 53.8 & 99.0 \\
Gemini 3.8 Flash\smash{$^\ast$} & 2026-09-02 & Gemini Flash & 94.2 & 67.3 & 55.8 & 63.5 & 25.0 & 40.4 & 57.7 & 90.1 \\
GPT-6-astra & 2026-09-03 & GPT flagship & 98.1 & 73.1 & 51.9 & 55.8 & 53.8 & 73.1 & 67.6 & 91.3 \\
Claude Opus 5.5\smash{$^\ast$} & 2026-09-22 & Claude Opus & 96.2 & 73.1 & 55.8 & 51.9 & 51.9 & 76.9 & 67.6 & 99.4 \\
GPT-6-luna\smash{$^\ast$} & 2026-09-22 & GPT small & 90.4 & 42.3 & 28.8 & 17.3 & 15.4 & 26.9 & 36.9 & 90.7 \\
Claude Sonnet 5.5\smash{$^\ast$} & 2026-09-28 & Claude Sonnet & 88.5 & 57.7 & 42.3 & 40.4 & 28.8 & 46.2 & 50.6 & 98.7 \\
GPT-6.1-sol\smash{$^\ast$} & 2026-09-29 & GPT flagship & 96.2 & 63.5 & 53.8 & 51.9 & 48.1 & 75.0 & 64.7 & 94.6 \\
Claude Haiku 5.5\smash{$^\ast$} & 2026-10-07 & Claude Haiku & 92.3 & 48.1 & 42.3 & 32.7 & 7.7 & 30.8 & 42.3 & 97.1 \\
Gemini 4 Argon\smash{$^\ast$} & 2026-10-13 & Gemini Pro & 94.2 & 61.5 & 34.6 & 42.3 & 38.5 & 51.9 & 53.8 & 96.2 \\
\bottomrule
\end{tabular}
\caption{The 44 models of the timeline analysis on the analysis subset (312 tasks), sorted by release date: all-pass rate (\%) by edit type and overall, and the percentage of edited programs that execute and render (Exec.). $^\ast$: one of the 22 models of Table~\ref{tab:main-results-full}. $^\dagger$: approximate release date. Claude Sonnet 4.6, Claude Sonnet 5, and GPT-5.6-luna are the medium levels of their effort sweeps.}
\label{tab:timeline-models}
\end{table*}

\paragraph{Models over time.}
Table~\ref{tab:timeline-models} lists the 44 models of Figure~\ref{fig:timeline}: the 22 models of Table~\ref{tab:main-results-full} and 22 earlier or intermediate releases of the same families. The best rate so far rises from 9.0\% (Qwen2.5-Coder-32B, November 2024) to 40.1\% (GPT-5.4, March 2026) and 67.6\% (GPT-6-astra, September 2026, matched by Claude Opus 5.5). Within the Claude Opus line, the rate rises from 30.4\% (Opus 4.5, November 2025) to 39.4\% (Opus 4.8, May 2026) and 67.6\% (Opus 5.5, September 2026). The best model available when Qwen3.5-27B (16.0\%) and Qwen3.8-27B (26.0\%) were released, Claude Sonnet 4.6 (36.2\%) and GPT-5.6-sol (53.2\%), led them by 20 and 27 points. Claude Sonnet 4, Sonnet 4.5, Opus 4.5, and Haiku 4.5 run with a thinking budget of 40k tokens instead of an effort level, and Gemini 2.5 Flash with 8,192.

\paragraph{Agreement and complementarity of the main results.}
Over the 17 models of Table~\ref{tab:main-results}, the six edit types give nearly the same ranking: the mean Spearman correlation between the per-type rankings is 0.97, and the lowest is 0.94. Table~\ref{tab:complementarity} compares the two strongest models task by task. Their passes are least correlated on S and C, where the tasks solved by at least one of them exceed the better single model by 17 and 15 points.

\begin{table}[t]
\centering
\small
\setlength{\tabcolsep}{4pt}
\begin{tabular}{@{}lrrrr@{}}
\toprule
Type & Opus 5.5 & GPT-6.1-sol & Either & $\phi$ \\
\midrule
G1 & 96.5 & 95.6 & 98.1 & 0.45 \\
G2 & 76.9 & 71.3 & 85.6 & 0.40 \\
G3 & 64.4 & 65.3 & 78.0 & 0.42 \\
G4 & 58.6 & 59.3 & 73.6 & 0.39 \\
S & 52.5 & 57.6 & 74.3 & 0.23 \\
C & 70.8 & 65.7 & 86.1 & 0.18 \\
\midrule
All & 69.9 & 69.1 & 82.6 & 0.38 \\
\bottomrule
\end{tabular}
\caption{\looseness=-1
The two strongest models on the full benchmark: all-pass rate (\%) of Claude Opus 5.5 and GPT-6.1-sol, the percentage of tasks solved by at least one of them (Either), and the task-level correlation of their passes ($\phi$).}
\label{tab:complementarity}
\end{table}

\begin{table}[!htb]
\centering
\footnotesize
\setlength{\tabcolsep}{3pt}
\begin{tabular}{@{}l p{0.57\columnwidth} r@{}}
\toprule
Code & Definition & Share \\
\midrule
\multicolumn{3}{@{}l}{\textit{Instruction: what to build}} \\
I-OMIT & A requested change is not attempted & 4.1\% \\
I-MISREAD & A different part, feature, or relation is built & 5.3\% \\
I-LITERAL & The wording is met but the intent is defeated & 0.3\% \\
I-OVERREACH & Rebuilt far beyond the request & 1.0\% \\
\midrule
\multicolumn{3}{@{}l}{\textit{Program: understanding the source}} \\
P-FRAME & A later transform re-orients or re-proportions the edit & 5.8\% \\
P-LOCATE & Edits code that does not produce the named feature & 2.7\% \\
P-COUPLE & Later code or dependent parameters undo or distort the edit & 8.5\% \\
P-CONTRACT & Uses an existing helper under a wrong assumption & 5.8\% \\
P-REWRITE & Rewrites a working helper unasked and breaks it & 6.3\% \\
P-OTHER & Other & 0.2\% \\
\midrule
\multicolumn{3}{@{}l}{\textit{Code: writing code that runs}} \\
K-API & Wrong or nonexistent Blender API call or argument & 6.2\% \\
K-LOGIC & Ordinary bug in new code & 11.1\% \\
K-SILENT & An operation runs without effect, e.g., a boolean no-op & 5.0\% \\
K-COST & Timeout or memory blow-up & 0.4\% \\
K-FORMAT & Reply not runnable: truncated or malformed & 5.6\% \\
\midrule
\multicolumn{3}{@{}l}{\textit{Geometry: foreseeing the built shape}} \\
G-TIMID & Right direction, too small & 3.5\% \\
G-OVER & Right direction, too large & 0.8\% \\
G-PARAM & A generator parameter changes the shape in an unintended way & 7.3\% \\
G-INTERACT & A new or changed part collides with, is swallowed by, or opens gaps in other parts & 9.2\% \\
G-SUPPORT & Parts float or detach & 5.3\% \\
G-GESTALT & Parts are built but the overall form is wrong & 3.8\% \\
G-OTHER & Other & 0.8\% \\
\midrule
\multicolumn{3}{@{}l}{\textit{Output: making the change visible}} \\
O-LEGIBLE & The change is built but too thin, hidden, or in the same material to be seen & 0.8\% \\
\bottomrule
\end{tabular}
\caption{Codebook of the error census. Each failure receives one stage, the earliest at which the edit diverges from the instruction, and one mechanism within it. Share: percentage of all failures of the 76 configurations on the analysis subset that receive the code.}
\label{tab:taxonomy-full}
\end{table}

\section{Editor Error Analysis}
\label{app:errors}

\begin{table*}[t]
\centering
\scriptsize
\setlength{\tabcolsep}{2.2pt}
\begin{tabular}[t]{@{}lrrrrrrr@{}}
\toprule
Configuration & Pass & I & P & K & G & O & Crash \\
\midrule
GPT-6.1-sol (xhigh) & 74.7\% & 7\% & 45\% & 19\% & 21\% & 7\% & 17 \\
Claude Sonnet 5.5 (xhigh) & 73.7\% & 13\% & 21\% & 21\% & 42\% & 3\% & 3 \\
Claude Opus 5.5 (xhigh) & 72.1\% & 0\% & 27\% & 17\% & 53\% & 3\% & 0 \\
GPT-6.1-sol (high) & 70.8\% & 6\% & 32\% & 28\% & 30\% & 4\% & 20 \\
GPT-6.1-sol (max) & 70.5\% & 4\% & 29\% & 29\% & 29\% & 10\% & 21 \\
Claude Opus 5.5 (high) & 68.9\% & 10\% & 18\% & 16\% & 49\% & 8\% & 2 \\
Claude Opus 5.5 & 67.6\% & 10\% & 30\% & 5\% & 50\% & 5\% & 2 \\
GPT-6-astra & 67.6\% & 5\% & 42\% & 31\% & 18\% & 5\% & 26 \\
GPT-6.1-sol (low) & 67.0\% & 7\% & 23\% & 34\% & 34\% & 2\% & 10 \\
Claude Sonnet 5.5 (high) & 65.4\% & 10\% & 24\% & 9\% & 54\% & 3\% & 3 \\
GPT-6.1-sol & 64.7\% & 7\% & 30\% & 28\% & 32\% & 3\% & 18 \\
Claude Opus 5.5 (low) & 59.9\% & 6\% & 25\% & 13\% & 52\% & 5\% & 0 \\
Gemini 3.8 Flash (max) & 58.7\% & 1\% & 26\% & 33\% & 38\% & 1\% & 19 \\
Gemini 3.8 Flash & 57.7\% & 3\% & 21\% & 39\% & 33\% & 4\% & 27 \\
GPT-5.6-luna (max) & 55.8\% & 13\% & 28\% & 16\% & 40\% & 3\% & 6 \\
Kimi K3 (max) & 54.5\% & 10\% & 29\% & 18\% & 43\% & 1\% & 8 \\
GPT-6-luna (max) & 54.2\% & 20\% & 19\% & 15\% & 44\% & 1\% & 7 \\
Claude Fable 5.1 & 53.8\% & 8\% & 19\% & 15\% & 53\% & 5\% & 1 \\
Gemini 4 Argon & 53.8\% & 8\% & 18\% & 24\% & 48\% & 2\% & 8 \\
Kimi K3 (xhigh) & 53.8\% & 10\% & 23\% & 21\% & 43\% & 3\% & 7 \\
GPT-5.6-sol & 53.2\% & 11\% & 20\% & 26\% & 42\% & 1\% & 14 \\
Claude Opus 5 & 52.2\% & 6\% & 22\% & 15\% & 55\% & 2\% & 3 \\
GPT-5.5 & 51.6\% & 13\% & 20\% & 20\% & 45\% & 2\% & 10 \\
GPT-6-luna (xhigh) & 51.6\% & 13\% & 25\% & 19\% & 38\% & 4\% & 21 \\
Claude Sonnet 5.5 & 50.6\% & 15\% & 22\% & 13\% & 47\% & 2\% & 4 \\
Gemini 3.8 Flash (high) & 50.3\% & 2\% & 21\% & 50\% & 26\% & 2\% & 33 \\
Kimi K3 (high) & 50.3\% & 11\% & 24\% & 18\% & 46\% & 1\% & 11 \\
Kimi K3 & 46.8\% & 8\% & 31\% & 22\% & 39\% & 1\% & 17 \\
GPT-5.6-luna (xhigh) & 45.5\% & 12\% & 26\% & 24\% & 36\% & 1\% & 13 \\
Gemini 3.7 Flash & 45.2\% & 4\% & 18\% & 35\% & 42\% & 1\% & 29 \\
Claude Sonnet 5.5 (low) & 42.9\% & 13\% & 25\% & 19\% & 43\% & 0\% & 5 \\
GPT-6-luna (high) & 42.9\% & 18\% & 21\% & 25\% & 34\% & 3\% & 28 \\
Claude Haiku 5.5 & 42.3\% & 14\% & 28\% & 12\% & 45\% & 1\% & 3 \\
GPT-5.6-luna (high) & 42.3\% & 10\% & 29\% & 25\% & 36\% & 1\% & 16 \\
Gemini 3.6 Flash & 41.3\% & 6\% & 26\% & 30\% & 38\% & 1\% & 24 \\
Gemini 3.8 Flash (low) & 41.0\% & 3\% & 20\% & 38\% & 38\% & 2\% & 26 \\
Claude Sonnet 5 (xhigh) & 40.1\% & 9\% & 26\% & 13\% & 50\% & 1\% & 7 \\
GPT-5.4 & 40.1\% & 14\% & 23\% & 15\% & 47\% & 1\% & 9 \\
\bottomrule
\end{tabular}
\hfill
\begin{tabular}[t]{@{}lrrrrrrr@{}}
\toprule
Configuration & Pass & I & P & K & G & O & Crash \\
\midrule
Claude Opus 4.8 & 39.4\% & 8\% & 22\% & 15\% & 53\% & 2\% & 8 \\
Gemini 3.5 Flash & 38.8\% & 9\% & 28\% & 26\% & 36\% & 1\% & 27 \\
GPT-6-luna & 36.9\% & 19\% & 23\% & 24\% & 34\% & 0\% & 24 \\
Claude Sonnet 4.6 & 36.2\% & 17\% & 19\% & 18\% & 47\% & 0\% & 12 \\
Claude Sonnet 4.6 (high) & 35.9\% & 7\% & 19\% & 32\% & 41\% & 1\% & 7 \\
Claude Sonnet 5 (high) & 35.3\% & 7\% & 27\% & 16\% & 49\% & 1\% & 9 \\
Gemini 3.1 Pro & 35.3\% & 10\% & 29\% & 13\% & 48\% & 1\% & 11 \\
Claude Opus 4.6 & 34.3\% & 7\% & 29\% & 24\% & 39\% & 1\% & 11 \\
Kimi K3 (low) & 30.8\% & 10\% & 33\% & 22\% & 35\% & 0\% & 13 \\
Claude Opus 4.5 & 30.4\% & 7\% & 25\% & 29\% & 39\% & 0\% & 13 \\
Claude Sonnet 4.6 (low) & 29.5\% & 14\% & 27\% & 21\% & 38\% & 0\% & 5 \\
Claude Sonnet 5 & 28.8\% & 13\% & 31\% & 18\% & 37\% & 0\% & 12 \\
GPT-6-luna (low) & 27.9\% & 21\% & 25\% & 22\% & 32\% & 0\% & 34 \\
Gemini 3 Flash & 27.9\% & 5\% & 35\% & 24\% & 36\% & 0\% & 35 \\
GPT-5.6-luna & 26.3\% & 14\% & 25\% & 37\% & 24\% & 0\% & 66 \\
Qwen3.8-27B & 26.0\% & 14\% & 33\% & 27\% & 26\% & 1\% & 30 \\
Kimi K3 (none) & 24.4\% & 18\% & 31\% & 22\% & 29\% & 0\% & 12 \\
Claude Sonnet 5 (low) & 24.0\% & 7\% & 36\% & 23\% & 34\% & 0\% & 9 \\
Claude Sonnet 4.5 & 20.5\% & 9\% & 39\% & 22\% & 30\% & 0\% & 24 \\
Qwen3.6-27B & 19.6\% & 9\% & 33\% & 32\% & 25\% & 0\% & 64 \\
Claude Haiku 4.5 & 18.6\% & 20\% & 29\% & 14\% & 37\% & 0\% & 30 \\
GPT-5.6-luna (low) & 18.3\% & 14\% & 22\% & 40\% & 24\% & 0\% & 73 \\
Claude Sonnet 4 & 17.3\% & 8\% & 29\% & 22\% & 40\% & 0\% & 33 \\
GPT-5.6-luna (none) & 16.0\% & 17\% & 37\% & 27\% & 18\% & 0\% & 64 \\
Gemma 4 31B & 16.0\% & 11\% & 35\% & 25\% & 28\% & 0\% & 36 \\
Qwen3.5-27B & 16.0\% & 8\% & 28\% & 45\% & 19\% & 0\% & 55 \\
Gemini 2.5 Flash & 15.7\% & 7\% & 33\% & 44\% & 15\% & 0\% & 115 \\
gpt-oss-120b & 14.4\% & 13\% & 38\% & 27\% & 22\% & 0\% & 67 \\
Qwen3.6-35B-A3B & 13.5\% & 10\% & 41\% & 29\% & 20\% & 0\% & 95 \\
Qwen3-VL-235B-A22B & 11.9\% & 10\% & 33\% & 36\% & 21\% & 0\% & 68 \\
Qwen3.5-35B-A3B & 11.5\% & 14\% & 32\% & 36\% & 18\% & 0\% & 85 \\
Qwen3-32B & 9.6\% & 12\% & 54\% & 19\% & 15\% & 0\% & 134 \\
Qwen3.5-4B & 9.6\% & 14\% & 26\% & 56\% & 4\% & 0\% & 120 \\
QwQ-32B & 9.0\% & 11\% & 57\% & 23\% & 10\% & 0\% & 156 \\
Qwen2.5-Coder-32B & 9.0\% & 12\% & 45\% & 29\% & 14\% & 0\% & 127 \\
Qwen3-Coder-Next & 8.7\% & 8\% & 34\% & 45\% & 13\% & 0\% & 136 \\
Qwen3.5-9B & 7.7\% & 10\% & 32\% & 49\% & 9\% & 0\% & 129 \\
Qwen3.5-2B & 0.3\% & 6\% & 5\% & 89\% & 0\% & 0\% & 17 \\
\bottomrule
\end{tabular}
\caption{All-pass rate (Pass) and stage mix of failures for all 76 configurations on the analysis subset, sorted by all-pass rate. Stages: I instruction, P program, K code, G geometric foresight, O visibility. Crash: number of tasks, out of 312, in which the edited program crashes. Effort levels other than medium are given in parentheses.}
\label{tab:census-profile}
\end{table*}

\paragraph{Coding procedure.}
We code every failure of the 76 single-shot configurations on the analysis subset: the 22 models of Table~\ref{tab:main-results-full}, 22 earlier and intermediate models, and 32 further effort levels. Failures that do not execute are classified from their error logs and crash sites with fixed rules. Rendered failures are coded by Gemini 3.8 Flash from the instruction, the checklist, both programs, and the twelve renders. The codebook (Table~\ref{tab:taxonomy-full}) gives each failure a stage, the first point at which the edit diverges from the instruction (I: instruction, P: program, K: code, G: geometric foresight, O: visibility), and one mechanism within that stage. Cases that cannot be attributed to a stage are left out. To check the coding, one of the authors independently labeled the stage of 150 randomly sampled rendered failures; agreement with the Flash coder is Cohen's $\kappa=0.71$.

\paragraph{Per-configuration profiles.}
Table~\ref{tab:census-profile} lists, for every configuration, the stage mix of its failures and how often its edited programs crash.

\end{document}